%% file: main.tex
\documentclass[letterpaper]{article} 
\usepackage[draft]{aaai25}  
\usepackage{times}  
\usepackage{helvet}  
\usepackage{courier}  
\usepackage[hyphens]{url}  
\usepackage{graphicx} 
\usepackage{natbib}  
\usepackage{caption} 
\usepackage{algorithm}
\usepackage{algorithmic}
\usepackage{xcolor}
\usepackage{amsmath}
\usepackage{booktabs}
\usepackage{multirow}
\usepackage{cleveref}
\usepackage{comment}
\usepackage{placeins}

\usepackage{newfloat}
\usepackage{listings}
\DeclareCaptionStyle{ruled}{labelfont=normalfont,labelsep=colon,strut=off} 
\floatstyle{ruled}
\newfloat{listing}{tb}{lst}{}
\floatname{listing}{Listing}
\title{IAO Prompting: Making Knowledge Flow Explicit in LLMs through Structured Reasoning Templates}
\author{
    A\"issatou Diallo$^1$\thanks{ Corresponding author: \texttt{a.diallo@ucl.ac.uk}\\}, Antonis Bikakis$^2$, Luke Dickens$^2$, Anthony Hunter$^1$, Rob Miller$^2$
}
\affiliations{

    \textsuperscript{\rm 1}Department of Computer Science \\
\textsuperscript{\rm 2}Department of Information Studies     \\

University College London, United Kingdom 
}

\usepackage{bibentry}

\begin{document}

\maketitle

\begin{abstract}
While Large Language Models (LLMs) demonstrate impressive reasoning capabilities, understanding and validating their knowledge utilization remains challenging. Chain-of-thought (CoT) prompting partially addresses this by revealing intermediate reasoning steps, but the knowledge flow and application remain implicit. We introduce IAO (Input-Action-Output) prompting, a structured template-based method that explicitly models how LLMs access and apply their knowledge during complex reasoning tasks. IAO decomposes problems into sequential steps, each clearly identifying the input knowledge being used, the action being performed, and the resulting output. This structured decomposition enables us to trace knowledge flow, verify factual consistency, and identify potential knowledge gaps or misapplications. Through experiments across diverse reasoning tasks, we demonstrate that IAO not only improves zero-shot performance but also provides transparency in how LLMs leverage their stored knowledge. Human evaluation confirms that this structured approach enhances our ability to verify knowledge utilization and detect potential hallucinations or reasoning errors. Our findings provide insights into both knowledge representation within LLMs and methods for more reliable knowledge application.
\end{abstract}

%

\section{Introduction}

\input{paper/intro}

\section{Related Work}

\input{paper/related_work}

\section{IAO Prompting}
\input{paper/approach}

\section{Experimental Setup}
\input{paper/experiments}

\section{Results}

\input{paper/results}

\section{Discussion}
\input{paper/discussion}

\section{Conclusion}

In this paper, we propose IAO prompting, a novel template-based approach that guides Large Language Models (LLMs) towards decomposing complex problems into manageable steps. By explicitly structuring input information, planned actions, and intermediate outputs, IAO facilitates sequential problem-solving. Our evaluations across diverse reasoning tasks demonstrate performance gains, increased clarity, improved structural coherence, and enhanced transparency within the reasoning chain. Moreover, IAO shows domain independence, requiring minimal adaptation to cater to different problem types. Additional human evaluation validates IAO utility in finding errors in free-text reasoning chains.

\section*{Acknowledgements}

This research was supported by the Leverhulme Trust grant for the project 'Repurposing of Resources: from Everyday Problem Solving through to Crisis Management' (RPG-2021-182). We also thank all reviewers for their insightful feedback.

\bibliography{aaai25}

\appendix

\section*{Limitations}
While IAO prompting demonstrates compelling advantages in terms of performance and transparency, it is crucial to acknowledge potential limitations:
\paragraph{Output Length}: The structured nature of IAO prompts might lead to lengthier responses compared to baseline models. This can have implications on computational costs and real-time applicability, particularly in resource-constrained settings. Future work could explore techniques for compressing the output or developing domain-specific adaptations to mitigate this limitation.
\paragraph{Interpretability Trade-off}: While the structured output enhances interpretability, it is important to consider that some users might prefer more concise summaries. The human evaluation study supports this. In fact, when the answer is correct, evaluators heavily preferred CoT reasoning chains over IAO's. This suggests a potential trade-off between detailed explanations and user preferences. Future work could explore methods to balance the level of detail and provide tailored interpretations based on individual user needs.

\section*{Ethical Considerations}

While IAO holds promise for improved reasoning, concerns arise regarding potential bias amplification and misuse as per any prompting method for LLMs. Breaking down complex tasks into smaller steps could inadvertently magnify existing biases in the LLM or training data, leading to biased final outputs. This necessitates careful bias detection and mitigation. Additionally, the structured nature could be exploited to "trick" the model, generating harmful or misleading outputs. Safeguards like fact-checking and verification become crucial, especially in sensitive domains. Responsible deployment and use are fundamental to ensure LLMs benefits are harnessed ethically and its risks are minimized.

\begin{table*}[ht]
\resizebox{\textwidth}{!}{%
\begin{tabular}{@{}llllll@{}}
\toprule
Prompt                                              & Date Understanding & Last Letter & AQuA & CommonsenseQA & \multicolumn{1}{c}{Average} \\ \midrule
\texttt{[Step, Input, Action, Output]}              & 82.4               & 46.0        & 64.6 & 82.7          & 68.9                        \\
\texttt{[Step, Subquestion, Action, Output]}         & 81.8               & 84.8        & 63.0 & 81.2          & 77.7                        \\
\texttt{[Step, Subquestion, Input, Output]}         & 85.9               & 76.0        & 61.0 & 82.5          & 76.3                        \\
\texttt{[Step, Subquestion, Input, Action]}         & 86.2               & 4.4         & 62.6 & 82.9          & 59.0                        \\ \midrule
\texttt{[Step, Subquestion, Input, Action, Output]} & 88.1               & 88.8        & 63.9 & 83.1          & 80.9                        \\ \bottomrule
\end{tabular}%
}
\caption{Performance in a zero-shot setting when one of the field of the prompt is removed.}
\label{tab:ablation_results_appendix}
\end{table*}

\section{Compute cost}
PaLM-2 (text-unicorn) has a cost of per API requests: \$0.0025/1K characters in input and \$0.0075/1K characters in output. GPT-4 has a cost \$0.01/1K tokens and \$0.03/1K tokens.

\section{Datasets}
\label{sec:dataset_appendix}

All included datasets are in English. We note that AQuA-RAT, Date Understanding, Object Shuffling are under the Apache License, Version 2.0. GSM8K and StrategyQA are under the MIT License. We also note that the datasets may include names of individuals collected from
the internet, i.e., publicly available facts about a
person but not in an offensive way. The following list shows the sources of data we used for this study:
\begin{itemize}
\small
    \item AQuA-RAT: \url{https://github.com/google-deepmind/AQuA}
    \item GSM8K: \url{https://github.com/openai/grade-school-math}
    \item StrategyQA: \url{https://github.com/google/BIGbench/tree/main/bigbench/benchmark_tasks/strategyqa}
    \item Last Letter: \url{https://github.com/kojima-takeshi188/zero_shot_cot/tree/main/dataset/last_letters}
    \item Date Understanding: from BIG-Bench (BIG-bench collaboration, 2021): \url{https://github.com/google/BIG-bench/blob/main/}
    \item Object Tracking: from BIG-Bench (BIG-bench collaboration, 2021): \url{https://github.com/google/BIG-bench/blob/main/}
    \item CommonsenseQA: \url{https://www.tau-nlp.sites.tau.ac.il/commonsenseqa}
\end{itemize}%

\begin{table}[H]
\centering
\resizebox{0.9\columnwidth}{!}{%
\begin{tabular}{@{}lll@{}}
\toprule
Dataset            & Avg words & Filename                \\ \midrule
AQuA               & 51.9      & test.jsonl              \\
GSM8k              & 46.9      & test.jsonl              \\
D.U. & 35.0      & task.json               \\
O.T.    & 91.1      & three\_objects/task.json \\
L.L       & 15.0      & last\_letters.json       \\
C.QA      & 27.8      & dev\_rand\_split.jsonl    \\
S.QA         & 9.6       & task.json               \\ \bottomrule
\end{tabular}%
}
\caption{Datasets statistics and files used. D.U: Date Understanding, O.T: Object Tracking, L.L: Last Letters Concatenation, C.QA: CommonsenseQA, S.QA: StrategyQA}
\label{tab:more_data_stats}
\end{table}

\begin{table}[H]
\resizebox{0.9\columnwidth}{!}{%
\begin{tabular}{@{}llll@{}}
\toprule
Reasoning Type                                  & Dataset            & Size & Answer Type     \\ \midrule
\multirow{2}{*}{\textit{Arithmetic Reasoning}}  & AQUA               & 254  & Multiple Choice \\
                                                & GSM8k              & 1319 & Numeral         \\ \midrule
\multirow{2}{*}{\textit{Commonsense Reasoning}} & StrategyQA         & 2290 & Yes/No          \\
                                                & CommonsenseQA      & 1221 & Multiple Choice \\ \midrule
\multirow{2}{*}{\textit{Other Reasoning}}       & Date Understanding & 369  & Multiple Choice \\
                                                & Object Tracking    & 750  & Multiple Choice \\ \midrule
\textit{Symbolic Reasoning}                     & Last Letter        & 500  & String          \\ \bottomrule
\end{tabular}%
}
\caption{Tasks, data statistics and answer type.}
\label{tab:data_statistics}
\end{table}

\begin{table}[H]
\centering
\resizebox{0.95\columnwidth}{!}{%
\begin{tabular}{@{}ll@{}}
\toprule
Dataset            & Answer prompt                               \\ \midrule
AQuA               & Therefore, among A through E, the answer is \\
GSM8k              & Therefore, the answer (arabic numerals) is  \\
D.U. & Therefore, among A through F, the answer is \\
O.T.    & Therefore, among A through C, the answer is \\
L.L.       & Therefore, the answer is                    \\
C.QA      & Therefore, among A through E, the answer is \\
S.QA         & The answer (Yes or No) is                   \\ \bottomrule
\end{tabular}%
}
\caption{Answer extraction prompts used. D.U: Date Understanding, O.T: Object Tracking, L.L: Last Letters Concatenation, C.QA: CommonsenseQA, S.QA: StrategyQA}
\label{tab:answer-prompt}
\end{table}

\section{Ablation studies}
\Cref{tab:ablation_results_appendix} shows how removing each field impacts the model's accuracy on different tasks. This helps understand the importance and relevance of each field in guiding the model towards successful completion of tasks.
For example, removing the "Subquestion" field seems to have a significant negative impact on the "Date Understanding" accuracy, suggesting that subquestions might provide crucial context for understanding temporal information.
The Last Letter task seems most affected by any field removal, highlighting its dependence on full prompt context.

\section{Experimental setting}
All results reported for PALM-2 are the results of average over 3 runs. The results reported for GPT-4 are over a single run due to higher costs. 
For each model, the temperature was set to 0, the maximum number of output tokens was set to 1024.
\begin{itemize}
    \item \texttt{gpt-4-1106-preview}: \url{https://platform.openai.com/docs/models/gpt-4-and-gpt-4-turbo}
    \item \texttt{PALM-2 (text-unicorn)}: \url{https://cloud.google.com/vertex-ai/docs/generative-ai/model-reference/text}
\end{itemize}

\section{Output word count}

\begin{table}[H]
\centering
\resizebox{0.4\columnwidth}{!}{%
\begin{tabular}{@{}lll@{}}
\toprule
                & \textbf{CoT} & \textbf{IAO} \\ \midrule
\textbf{PALM-2} & 144.3        & 155.2        \\
\textbf{GPT-4}  & 188.2        & 175.4        \\ \bottomrule
\end{tabular}%
}
\caption{Average output word count.}\label{tab:wc_avg}
\end{table}

We sampled 50 reasoning chains for each model across various benchmarks and report the average output word count.
GPT-4 tends to be more verbose in CoT and IAO structure helps in shortening the output reasoning but the opposite is seen in PALM-2.

\section{Additional Examples}

\begin{table*}[h!]
\resizebox{\textwidth}{!}{%
\begin{tabular}{@{}llllll@{}}
\toprule
\multicolumn{2}{l}{Question} &
  \multicolumn{4}{l}{Take the last letters of each words in "Pauline Kerry Jeannette Hope" and concatenate them. \hspace{5cm} \textcolor{blue}{(GT: "eyee")}} \\ \midrule
\multirow{4}{*}{\rotatebox[origin=c]{90}{PALM-2}} &
  CoT &
  \multicolumn{4}{l}{\begin{tabular}[c]{@{}l@{}}The last letters of each word in "Pauline Kerry Jeannette Hope" are "e", "y", "e", and "e".\\ Concatenating them, we get "eyeee".\\ Therefore, the answer is \textbf{eyeee}.\end{tabular}} \\ \cmidrule(l){2-6} 
 &
  \multirow{3}{*}{IAO} &
  Subquestion &
  Input &
  Action &
  Output \\ \cmidrule(l){3-6} 
 &
   &
  \begin{tabular}[c]{@{}l@{}}1. What are the last letters of each word in \\ "Pauline Kerry Jeannette Hope"?\end{tabular} &
  "Pauline Kerry Jeannette Hope" &
  Get the last letters of each word &
  "e y e e" \\
 &
   &
  2. Concatenate the last letters of each word. &
  "e y e e" &
  Concatenate the letters. &
  \textbf{"eyee"} \\ \midrule
\multirow{7}{*}{\rotatebox[origin=c]{90}{GPT-4}} &
  CoT &
  \multicolumn{4}{l}{\begin{tabular}[c]{@{}l@{}}To solve this, we need to take the last letter of each word in the phrase "Pauline Kerry Jeannette Hope" and then concatenate them, \\ which means to link them together in a sequence.\\ Here are the last letters of each word:\\ - Pauline: e\\ - Kerry: y\\ - Jeannette: e\\ - Hope: e\\ Now, we concatenate these letters: e + y + e + e\\ Therefore, the answer is \textbf{"eyee”}\end{tabular}} \\ \cmidrule(l){2-6} 
 &
  \multirow{6}{*}{IAO} &
  Subquestion &
  Input &
  Action &
  Output \\ \cmidrule(l){3-6} 
 &
   &
  1. What is the last letter of the first word "Pauline"? &
  The word "Pauline" &
  \begin{tabular}[c]{@{}l@{}}Identify the last letter of the \\ word "Pauline".\end{tabular} &
  The last letter is "e". \\
 &
   &
  2. What is the last letter of the second word "Kerry"? &
  The word "Kerry" &
  \begin{tabular}[c]{@{}l@{}}Identify the last letter of the \\ word "Kerry".\end{tabular} &
  The last letter is "y". \\
 &
   &
  3. What is the last letter of the third word "Jeannette"? &
  The word "Jeannette" &
  \begin{tabular}[c]{@{}l@{}}Identify the last letter of the \\ word "Jeannette".\end{tabular} &
  The last letter is "e". \\
 &
   &
  4. What is the last letter of the fourth word "Hope"? &
  The word "Hope" &
  \begin{tabular}[c]{@{}l@{}}Identify the last letter of the \\ word "Hope".\end{tabular} &
  The last letter is "e". \\
 &
   &
  3. Concatenate the last letters of each word. &
  The last letters "e", "y", "e", "e" &
  \begin{tabular}[c]{@{}l@{}}Concatenate the letters "e", "y", "e", "e" \\ in the order they were obtained.\end{tabular} &
  \begin{tabular}[c]{@{}l@{}}The concatenated result is \\ \textbf{"eyee"}.\end{tabular} \\ \bottomrule
\end{tabular}%
}
\caption{Output examples from Last Letters dataset. The final results are in \textbf{bold}.}
\label{tab:letter_example}
\end{table*}

\begin{table*}[h!]
\resizebox{\textwidth}{!}{%
\begin{tabular}{@{}llllll@{}}
\toprule
\multicolumn{2}{l}{Question} &
  \multicolumn{4}{l}{\begin{tabular}[c]{@{}l@{}}Today's meeting is rescheduled to 11 am tomorrow, 10/16/1924. What is the date 10 days ago in MM/DD/YYYY? \\ Answer Choices: (A) 10/05/1924 (B) 05/05/1924 (C) 10/12/1924 (D) 10/06/1924 (E) 11/02/1924 (F) 10/04/1924  \hspace{5cm}    \textcolor{blue}{(GT: (A))}\end{tabular}} \\ \midrule
\multirow{4}{*}{\rotatebox[origin=c]{90}{PALM-2}} &
  CoT &
  \multicolumn{4}{l}{\begin{tabular}[c]{@{}l@{}}The date of the meeting is 10/16/1924.\\ 10 days ago from 10/16/1924 is 10/06/1924.\\ Therefore, among A through F, the answer is \textbf{D}\end{tabular}} \\ \cmidrule(l){2-6} 
 &
  \multirow{3}{*}{IAO} &
  Subquestion &
  Input &
  Action &
  Output \\ \cmidrule(l){3-6} 
 &
   &
  1. What is the date 10 days ago? &
  10/16/1924 &
  subtract 10 days from the given date &
  10/06/1924 \\
 &
   &
  \begin{tabular}[c]{@{}l@{}}2. What is the answer choice that \\ matches the date 10 days ago?\end{tabular} &
  10/06/1924 &
  \begin{tabular}[c]{@{}l@{}}compare the date 10 days ago \\ with the answer choices\end{tabular} &
  \textbf{(D)} 10/06/1924 \\ \midrule
\multirow{5}{*}{\rotatebox[origin=c]{90}{GPT-4}} &
  CoT &
  \multicolumn{4}{l}{\begin{tabular}[c]{@{}l@{}}To find the date 10 days before 10/16/1924, we need to subtract 10 days from that date.\\ Starting with 10/16/1924:\\ 1. If we subtract 10 days from the 16th, we would be in the previous month, October, since October has 31 days.\\ 2. To find the exact date, we subtract 10 from 16, which gives us 6.\\ Therefore, the date 10 days before 10/16/1924 is 10/06/1924.\\ Among the answer choices provided:\\ (A) 10/05/1924\\ (B) 05/05/1924\\ (C) 10/12/1924\\ (D) 10/06/1924\\ (E) 11/02/1924\\ (F) 10/04/1924\\ The correct answer is \textbf{(D)} 10/06/1924.\end{tabular}} \\ \cmidrule(l){2-6} 
 &
  \multirow{4}{*}{IAO} &
  Subquestion &
  Input &
  Action &
  Output \\ \cmidrule(l){3-6} 
 &
   &
  \begin{tabular}[c]{@{}l@{}}1. What is the date 10 days \\ before 10/16/1924?\end{tabular} &
  The given date is 10/16/1924. &
  Subtract 10 days from the given date. &
  \begin{tabular}[c]{@{}l@{}}The date 10 days before\\ 10/16/1924.\end{tabular} \\
 &
   &
  \multicolumn{4}{l}{{\begin{tabular}[c]{@{}l@{}}To find the date 10 days before 10/16/1924, we need to count backwards day by day. Since October has 31 days, subtracting 10 days\\  from the 16th will not take us into the previous month. Therefore, we simply subtract 10 from 16.\\ 16 - 10 = 6\\ The month (October) and the year (1924) remain the same. Therefore, the date 10 days before 10/16/1924 is 10/06/1924.\\ Therefore, among A through F, the answer is \textbf{(D)} 10/06/1924\end{tabular}}} \\
 &
   &
  \multicolumn{4}{l}{} \\ \bottomrule
\end{tabular}%
}
\caption{Output examples from Date Understanding dataset. The final results are in \textbf{bold}.}
\label{tab:date_example}
\end{table*}
\begin{table*}[h!]
\resizebox{\textwidth}{!}{%
\begin{tabular}{@{}llllll@{}}
\toprule
\multicolumn{2}{l}{Question} &
  \multicolumn{4}{l}{\begin{tabular}[c]{@{}l@{}}Alice, Bob, and Claire are friends and avid readers who occasionally trade books. At the start of the semester, they each buy one new book: Alice gets \\ The Great Gatsby, Bob gets The Odyssey, and Claire gets Lolita. As the semester proceeds, they start trading around the new books. First, Alice and \\ Bob swap books. Then, Alice and Claire swap books. Finally, Bob and Claire swap books. At the end of the semester, Bob has \\ Which choice is true ? Answer Choices: (A) The Great Gatsby. (B) The Odyssey. (C) Lolita. \hspace{4cm} \textcolor{blue}{GT: (B) The Odissey}\end{tabular}} \\ \midrule
\multirow{4}{*}{\rotatebox[origin=c]{90}{PALM-2}} &
  CoT &
  \multicolumn{4}{l}{\begin{tabular}[c]{@{}l@{}}At the start of the semester, Alice has The Great Gatsby, Bob has The Odyssey, and Claire has Lolita.\\ After Alice and Bob swap books, Alice has The Odyssey and Bob has The Great Gatsby.\\ After Alice and Claire swap books, Alice has Lolita and Claire has The Odyssey.\\ After Bob and Claire swap books, Bob has Lolita and Claire has The Great Gatsby.\\ Therefore, among A through C, the answer is \textbf{C}.\end{tabular}} \\ \cmidrule(l){2-6} 
 &
  \multirow{3}{*}{IAO} &
  Subquestion &
  Input &
  Action &
  Output \\ \cmidrule(l){3-6} 
 &
   &
  1. What book does Bob have after the first swap? &
  \begin{tabular}[c]{@{}l@{}}Alice has The Great Gatsby, \\ Bob has The Odyssey.\end{tabular} &
  Bob and Alice swap books. &
  Bob has The Great Gatsby. \\
 &
   &
  2. What book does Bob have after the second swap? &
  \begin{tabular}[c]{@{}l@{}}Bob has The Great Gatsby, \\ Claire has Lolita.\end{tabular} &
  Bob and Claire swap books. &
  Bob has The Great Gatsby. \\
 &
   &
  3. What book does Bob have at the end of the semester? &
  Bob has The Great Gatsby. &
  No more swaps. &
  Bob has \textbf{The Great Gatsby}. \\ \midrule
\multirow{5}{*}{\rotatebox[origin=c]{90}{GPT-4}} &
  CoT &
  \multicolumn{4}{l}{\begin{tabular}[c]{@{}l@{}}Let's track the movement of the books step by step:\\ 1. Initially:\\    - Alice has The Great Gatsby.\\    - Bob has The Odyssey.\\    - Claire has Lolita.\\ 2. After Alice and Bob swap books:\\    - Alice has The Odyssey.\\    - Bob has The Great Gatsby.\\    - Claire still has Lolita.\\ 3. Then Alice and Claire swap books:\\    - Alice has Lolita.\\    - Bob still has The Great Gatsby.\\    - Claire has The Odyssey.\\ 4. Finally, Bob and Claire swap books:\\    - Alice still has Lolita.\\    - Bob has The Odyssey.\\    - Claire has The Great Gatsby.\\ At the end of the semester, Bob has The Odyssey.\\ Therefore, among A through C, the answer is \textbf{(B) The Odyssey}.\end{tabular}} \\ \cmidrule(l){2-6} 
 &
  IAO &
  Subquestion &
  Input &
  Action &
  Output \\ \cmidrule(l){2-6} 
 &
   &
  1. What book does Bob have after he and Alice swap books? &
  \begin{tabular}[c]{@{}l@{}}Alice has The Great Gatsby, \\ Bob has The Odyssey.\end{tabular} &
  \begin{tabular}[c]{@{}l@{}}Swap the books between \\ Alice and Bob.\end{tabular} &
  \begin{tabular}[c]{@{}l@{}}Alice now has The Odyssey, \\ Bob now has The Great Gatsby.\end{tabular} \\
 &
   &
  2. What book does Alice have after she and Claire swap books? &
  \begin{tabular}[c]{@{}l@{}}Alice has The Odyssey, \\ Claire has Lolita.\end{tabular} &
  \begin{tabular}[c]{@{}l@{}}Swap the books between \\ Alice and Claire.\end{tabular} &
  \begin{tabular}[c]{@{}l@{}}Alice now has Lolita, \\ Claire now has The Odyssey.\end{tabular} \\
 &
   &
  3. What book does Bob have after he and Claire swap books? &
  \begin{tabular}[c]{@{}l@{}}Bob has The Great Gatsby, \\ Claire has The Odyssey.\end{tabular} &
  \begin{tabular}[c]{@{}l@{}}Swap the books between \\ Bob and Claire.\end{tabular} &
  \begin{tabular}[c]{@{}l@{}}Bob now has \textbf{The Odyssey}, \\ Claire now has The Great Gatsby.\end{tabular} \\ \bottomrule
\end{tabular}%
}
\caption{Output examples from Object Tracking dataset. The final results are in \textbf{bold}.}
\label{tab:object_example}
\end{table*}
\begin{table*}[h!]
\resizebox{\textwidth}{!}{%
\begin{tabular}{@{}llllll@{}}
\toprule
\multicolumn{2}{l}{Question} &
  \multicolumn{4}{l}{\begin{tabular}[c]{@{}l@{}}Andrea has 8 more apples than Jamal and half as many bananas as Jamal. Jamal has 4 more bananas than apples. How many fruits are there \\ if Andrea has 52 apples? \hspace{5cm} \textcolor{blue}{(GT: 168)}\end{tabular}} \\ \midrule
\multirow{6}{*}{\rotatebox[origin=c]{90}{PALM-2}} &
  CoT &
  \multicolumn{4}{l}{\begin{tabular}[c]{@{}l@{}}Let's denote the number of apples Jamal has as A and the number of bananas Jamal has as B.\\ From the given information, we know that Andrea has 8 more apples than Jamal, so Andrea has A + 8 apples. We also know that Andrea has half as \\ many bananas as Jamal, so Andrea has B/2 bananas.\\ We are given that Andrea has 52 apples, so we can write the equation A + 8 = 52. Solving for A, we get A = 44.\\ We are also given that Jamal has 4 more bananas than apples, so we can write the equation B = A + 4. \\ Substituting the value of A, we get B = 44 + 4 = 48.\\ Now we can calculate the total number of fruits. Andrea has A + 8 apples and B/2 bananas, so she has 44 + 8 + 48/2 = 72 fruits. \\ Jamal has A apples and B bananas, so he has 44 + 48 = 92 fruits.\\ Therefore, the total number of fruits is 72 + 92 = 164.\\ \\ Therefore, the answer is \textbf{164}.\end{tabular}} \\ \cmidrule(l){2-6} 
 &
  \multirow{5}{*}{IAO} &
  Subquestion &
  Input &
  Action &
  Output \\ \cmidrule(l){3-6} 
 &
   &
  1. How many apples does Jamal have? &
  \begin{tabular}[c]{@{}l@{}}Andrea has 8 more apples than Jamal.\\  Andrea has 52 apples.\end{tabular} &
  Jamal has 52 - 8 = 44 apples. &
  Jamal has 44 apples. \\
 &
   &
  2. How many bananas does Jamal have? &
  \begin{tabular}[c]{@{}l@{}}Jamal has 4 more bananas than apples. \\ Jamal has 44 apples.\end{tabular} &
  Jamal has 44 + 4 = 48 bananas. &
  Jamal has 48 bananas. \\
 &
   &
  3. How many bananas does Andrea have? &
  \begin{tabular}[c]{@{}l@{}}Andrea has half as many bananas as Jamal. \\ Jamal has 48 bananas.\end{tabular} &
  Andrea has 48 / 2 = 24 bananas. &
  Andrea has 24 bananas. \\
 &
   &
  4. How many fruits are there in total? &
  \begin{tabular}[c]{@{}l@{}}Andrea has 52 apples and 24 bananas. \\ Jamal has 44 apples and 48 bananas.\end{tabular} &
  There are 52 + 24 + 44 + 48 = 168 fruits in total. &
  There are \textbf{168} fruits in total. \\ \midrule
\multirow{8}{*}{\rotatebox[origin=c]{90}{GPT-4}} &
  CoT &
  \multicolumn{4}{l}{\begin{tabular}[c]{@{}l@{}}First, let's determine the number of apples and bananas Jamal has.\\ According to the information given, Andrea has 8 more apples than Jamal. Since Andrea has 52 apples, we can calculate the number of apples \\ Jamal has by subtracting 8 from Andrea's total:\\ Jamal's apples = Andrea's apples - 8\\ Jamal's apples = 52 - 8\\ Jamal's apples = 44\\ Now we know Jamal has 44 apples.\\ Next, we are told that Jamal has 4 more bananas than apples. Since we know Jamal has 44 apples, we can find out how many bananas he has:\\ Jamal's bananas = Jamal's apples + 4\\ Jamal's bananas = 44 + 4\\ Jamal's bananas = 48\\ \\ So, Jamal has 48 bananas.\\ Now, let's find out how many bananas Andrea has. We are told that Andrea has half as many bananas as Jamal. So we can calculate this as follows:\\ \\ Andrea's bananas = Jamal's bananas / 2\\ Andrea's bananas = 48 / 2\\ Andrea's bananas = 24\\ \\ Now we can calculate the total number of fruits by adding the number of apples and bananas both Andrea and Jamal have:\\ Total fruits = Andrea's apples + Andrea's bananas + Jamal's apples + Jamal's bananas\\ Total fruits = 52 + 24 + 44 + 48\\ Total fruits = 76 + 92\\ Total fruits = 168\\ \\ Therefore, the answer is 168.\end{tabular}} \\ \cmidrule(l){2-6} 
 &
  \multirow{7}{*}{IAO} &
  Subquestion &
  Input &
  Action &
  Output \\ \cmidrule(l){3-6} 
 &
   &
  How many apples does Jamal have? &
  \begin{tabular}[c]{@{}l@{}}Andrea has 8 more apples than Jamal, \\ Andrea has 52 apples.\end{tabular} &
  \begin{tabular}[c]{@{}l@{}}Subtract 8 from the number of apples \\ Andrea has to find out how many apples Jamal has.\end{tabular} &
  Jamal has 52 - 8 = 44 apples. \\
 &
   &
  How many bananas does Jamal have? &
  \begin{tabular}[c]{@{}l@{}}Jamal has 4 more bananas than apples, \\ Jamal has 44 apples.\end{tabular} &
  \begin{tabular}[c]{@{}l@{}}Add 4 to the number of apples Jamal has to find \\ out how many bananas he has.\end{tabular} &
  Jamal has 44 + 4 = 48 bananas. \\
 &
   &
  How many bananas does Andrea have? &
  \begin{tabular}[c]{@{}l@{}}Andrea has half as many bananas as Jamal, \\ Jamal has 48 bananas.\end{tabular} &
  \begin{tabular}[c]{@{}l@{}}Divide the number of bananas Jamal has by 2 to \\ find out how many bananas Andrea has.\end{tabular} &
  Andrea has 48 / 2 = 24 bananas. \\
 &
   &
  How many fruits does Andrea have in total? &
  Andrea has 52 apples and 24 bananas. &
  Add the number of apples and bananas Andrea has. &
  Andrea has 52 + 24 = 76 fruits. \\
 &
   &
  How many fruits does Jamal have in total? &
  Jamal has 44 apples and 48 bananas. &
  Add the number of apples and bananas Jamal has. &
  Jamal has 44 + 48 = 92 fruits. \\
 &
   &
  How many fruits are there in total? &
  Andrea has 76 fruits, and Jamal has 92 fruits. &
  \begin{tabular}[c]{@{}l@{}}Add the number of fruits Andrea has to the\\  number of fruits Jamal has.\end{tabular} &
  There are 76 + 92 = \textbf{168} fruits in total. \\ \bottomrule
\end{tabular}%
}
\caption{Output examples from GSM8k dataset. The final results are in \textbf{bold}.}
\label{tab:gsm8k_example}
\end{table*}

\begin{table*}[h]
\resizebox{\textwidth}{!}{%
\begin{tabular}{@{}llllll@{}}
\toprule
\multicolumn{2}{l}{Question} &
  \multicolumn{4}{l}{\begin{tabular}[c]{@{}l@{}}The man tried to reply to the woman, but he had difficulty keeping track of conversations that he didn't do what to? \\ Answer Choices: (A) initiate (B) ignore (C) question (D) answer (E) ask \hspace{5cm} \textcolor{blue}{(GT: (A)}\end{tabular}} \\ \midrule
\multirow{5}{*}{\rotatebox[origin=c]{90}{PALM-2}} &
  CoT &
  \multicolumn{4}{l}{\begin{tabular}[c]{@{}l@{}}The man tried to reply to the woman, but he had difficulty keeping track of conversations that he didn't do what to?\\ The man tried to reply to the woman, but he had difficulty keeping track of conversations. This means that he was not able to follow the conversation properly. \\ He was not able to understand what the woman was saying. This could be because he was not paying attention or because he was not able to \\ understand the language.\\ The question asks what the man did not do. \\ The answer choices are: (A) initiate (B) ignore (C) question (D) answer (E) ask.\\ The man did not initiate the conversation. He was replying to the woman. He did not ignore the woman. He was trying to reply to her. He did not question\\  the woman. He was trying to understand what she was saying. He did not answer the woman. He was trying to reply to her. He did not ask the woman anything. \\ He was trying to understand what she was saying.\\ \\ Therefore, among A through E, the answer is \textbf{(D)}\end{tabular}} \\ \cmidrule(l){2-6} 
 &
  \multirow{4}{*}{IAO} &
  Subquestion &
  Input &
  Action &
  Output \\ \cmidrule(l){3-6} 
 &
   &
  \begin{tabular}[c]{@{}l@{}}1. What is the main idea \\ of the question?\end{tabular} &
  \begin{tabular}[c]{@{}l@{}}The man tried to reply to the \\ woman, but he had difficulty\\ keeping track of conversations\\  that he didn't do what to?\end{tabular} &
  \begin{tabular}[c]{@{}l@{}}Identify the main idea \\ of the question.\end{tabular} &
  \begin{tabular}[c]{@{}l@{}}The main idea of the question is that \\ the man had difficulty keeping track \\ of conversations that he didn't do something to.\end{tabular} \\
 &
   &
  \begin{tabular}[c]{@{}l@{}}2. What is the missing \\ word in the question?\end{tabular} &
  \begin{tabular}[c]{@{}l@{}}The man tried to reply to the\\ woman, but he had difficulty \\ keeping track of conversations \\ that he didn't do what to?\end{tabular} &
  \begin{tabular}[c]{@{}l@{}}Identify the missing \\ word in the question.\end{tabular} &
  The missing word in the question is "initiate". \\
 &
   &
  \begin{tabular}[c]{@{}l@{}}3. What is the answer to\\  the question?\end{tabular} &
  \begin{tabular}[c]{@{}l@{}}The man tried to reply to the\\ woman, but he had difficulty \\ keeping track of conversations\\ that he didn't do what to?\end{tabular} &
  \begin{tabular}[c]{@{}l@{}}Choose the answer that \\ best fits the question.\end{tabular} &
  The answer to the question is \textbf{(A)} initiate. \\ \midrule
\multirow{6}{*}{\rotatebox[origin=c]{90}{GPT-4}} &
  CoT &
  \multicolumn{4}{l}{\begin{tabular}[c]{@{}l@{}}To solve this reasoning problem, we need to determine which action would make it difficult for the man to keep track of conversations. If he \\ has difficulty keeping track of conversations, it is likely because he is not actively engaged in starting or controlling the flow of the conversation.\\ \\ Let's look at the options:\\ \\ (A) initiate - If he doesn't initiate the conversation, he might have trouble keeping track because he is not setting the topic or pace of the conversation.\\ (B) ignore - Ignoring a conversation would not cause difficulty in keeping track; it would mean he is not participating at all.\\ (C) question - Not questioning might make it harder to understand, but it doesn't necessarily mean he would lose track of the conversation.\\ (D) answer - If he doesn't answer, it could be a result of losing track, not a cause.\\ (E) ask - Similar to initiate, if he doesn't ask questions, he might not be engaged enough to follow the conversation.\\ \\ Between initiating and asking, initiating a conversation is a more fundamental action for controlling the flow of the conversation. \\ If he doesn't initiate conversations, he is more likely to be a passive participant and thus have difficulty keeping track.\\ \\ Therefore, among A through E, the answer is \textbf{(A)} initiate\end{tabular}} \\ \cmidrule(l){2-6} 
 &
  \multirow{5}{*}{IAO} &
  Subquestion &
  Input &
  Action &
  Output \\ \cmidrule(l){3-6} 
 &
   &
  \begin{tabular}[c]{@{}l@{}}1. What does the phrase \\ "difficulty keeping track of \\ conversations" imply about \\ the man's ability to engage \\ in conversations?\end{tabular} &
  \begin{tabular}[c]{@{}l@{}}The man had difficulty keeping \\ track of conversations.\end{tabular} &
  \begin{tabular}[c]{@{}l@{}}Analyze the phrase to determine \\ what aspect of conversation the\\  man struggles with.\end{tabular} &
  \begin{tabular}[c]{@{}l@{}}The man struggles with following along\\  or maintaining focus in conversations \\ that he is not actively leading or starting.\end{tabular} \\
 &
   &
  \begin{tabular}[c]{@{}l@{}}2. Which answer choice is \\ related to leading or starting a\\  conversation?\end{tabular} &
  \begin{tabular}[c]{@{}l@{}}Answer choices (A) initiate \\ (B) ignore (C) question \\ (D) answer (E) ask\end{tabular} &
  \begin{tabular}[c]{@{}l@{}}Identify the answer choice that \\ means to begin or lead a \\ conversation.\end{tabular} &
  (A) initiate \\
 &
   &
  \begin{tabular}[c]{@{}l@{}}3. Does the difficulty in keeping \\ track of conversations imply that \\ the man did not do what?\end{tabular} &
  \begin{tabular}[c]{@{}l@{}}The man had difficulty keeping\\ track of conversations; the man \\ tried to reply to the woman.\end{tabular} &
  \begin{tabular}[c]{@{}l@{}}Determine which action from \\ the answer choices would be most\\  difficult for someone who cannot \\ keep track of conversations.\end{tabular} &
  \begin{tabular}[c]{@{}l@{}}The man would have difficulty replying \\ (or keeping track) if he did not initiate \\ the conversation.\end{tabular} \\
 &
   &
  \multicolumn{4}{l}{Therefore, among A through E, the answer is \textbf{(A)} initiate.}
\end{tabular}%
}
\caption{Output examples from CommonsenseQA dataset. The final results are in \textbf{bold}.}
\label{tab:commonsenseqa_example}
\end{table*}

\clearpage\clearpage

\section{Errors from AQua}
\textbf{Note:} Several questions where IAO gave the wrong answers were due to the output exceeding the maximum number of tokens. 
\subsection{Example 1}
\textit{Question}: A grocer makes a 25\% profit on the selling price for each bag of flour it sells. If he sells each bag for \$100 and makes \$3,000 in profit, how many bags did he sell? 

\textit{Answer Choices}: (A) 12 (B) 16 (C) 24 (D) 30 (E) 40"

\textit{GT}:  "C"

\paragraph{IAO}:
\noindent
\textbf{Step 1:}
\begin{itemize}
    \item \textit{Subquestion:} What is the profit made on one bag of flour?
    \item \textit{Input:} 25\% profit on the selling price of \$100 per bag
    \item \textit{Action:} Calculate the profit per bag by multiplying the selling price by the profit percentage.
    \item \textit{Output:} \$25 profit per bag (since 25\% of \$100 is \$25)
\end{itemize}

\textbf{Step 2:}
\begin{itemize}
    \item \textit{Subquestion:} How many bags of flour were sold to make \$3,000 in profit?
    \item \textit{Input:} \$3,000 total profit, \$25 profit per bag
    \item \textit{Action:} Divide the total profit by the profit per bag to find the number of bags sold.
    \item \textit{Output:} 120 bags sold (since \$3,000 / \$25 per bag = 120 bags)
\end{itemize}

\textbf{Step 3:}
\begin{itemize}
    \item \textit{Subquestion:} Which answer choice corresponds to the number of bags sold?
    \item \textit{Input:} 120 bags sold, Answer Choices: (A) 12 (B) 16 (C) 24 (D) 30 (E) 40
    \item \textit{Action:} Identify the correct answer choice that matches the number of bags sold.
    \item \textit{Output:} None of the answer choices match the number of bags sold.
\end{itemize}

Therefore, among answer choices A through E, the answer is none of the provided options. There seems to be a mistake as none of the answer choices correspond to the correct number of bags sold based on the given information.

\paragraph{CoT}: To solve this problem, we need to determine how much profit the grocer makes on each bag of flour and then use that information to find out how many bags were sold to make a total profit of \$3,000.

\textbf{Step 1: Calculate the profit per bag.} \\
Since the grocer makes a 25\% profit on the selling price, we can calculate the profit per bag as follows:

\begin{align*}
\text{Profit per bag} &= 25\% \text{ of selling price} \\
\text{Profit per bag} &= 0.25 \times \$100 \\
\text{Profit per bag} &= \$25
\end{align*}

\textbf{Step 2: Calculate the number of bags sold.} \\
Now that we know the profit per bag is \$25, we can calculate the number of bags sold to make a total profit of \$3,000.

\begin{align*}
\text{Total profit} &= \text{Profit per bag} \times \text{Number of bags sold} \\
\$3,000 &= \$25 \times \text{Number of bags sold}
\end{align*}

To find the number of bags sold, we divide the total profit by the profit per bag:

\begin{align*}
\text{Number of bags sold} &= \frac{\$3,000}{\$25} \\
\text{Number of bags sold} &= 120
\end{align*}

However, 120 is not one of the answer choices provided, which means we need to re-evaluate our calculations.

Let's go back to the statement that the grocer makes a 25\% profit on the selling price. This means that the selling price includes both the cost price and the profit. If the selling price is \$100 and the profit is 25\% of this selling price, then the profit per bag is indeed \$25, as we calculated.

Now, let's re-calculate the number of bags sold using the correct profit per bag:

\begin{align*}
\text{Number of bags sold} &= \frac{\$3,000}{\$25} \\
\text{Number of bags sold} &= 120
\end{align*}

Since 120 is not among the answer choices, we need to check if we've made a mistake in interpreting the answer choices or the question itself. If the question and calculations are correct, then there might be an error in the answer choices provided. If this is the case, none of the answer choices (A) through (E) is correct.

Let's review the question one more time to ensure we haven't missed any details. The question states that the grocer makes a 25\% profit on the selling price for each bag of flour and sells each bag for \$100. Our calculation of the profit per bag at \$25 is correct, and so is our calculation of the number of bags sold to make a \$3,000 profit.

Therefore, it seems there is a discrepancy with the answer choices provided, as none of them match our calculation of 120 bags. If we must choose from the given options, we would have to conclude that none of them is correct. However, since this is not a standard outcome for such problems, it's possible that there is a misunderstanding or an error in the question or answer choices.

\subsection{Example 2}

\textit{Question}: Find 25/12*5 

\textit{Answer Choices}: (A) 2.5498 (B) 0.4167 (C) 3.3987 (D) 8.5497 (E) 5.6312

\textit{GT}: B

\paragraph{IAO}: \noindent
\textbf{Step 1:}
\begin{itemize}
    \item \textit{Subquestion:} What is the product of $(\frac{25}{12})$ and 5?
    \item \textit{Input:} $(\frac{25}{12})$, 5
    \item \textit{Action:} Multiply $(\frac{25}{12})$ by 5.
    \item \textit{Output:} 
    $\frac{25}{12} \times 5 = \frac{25 \times 5}{12} = \frac{125}{12}$
\end{itemize}
\textbf{Step 2:}
\begin{itemize}
    \item \textit{Subquestion:} What is the decimal equivalent of 125/12?
    \item \textit{Input:} 125/12
    \item \textit{Action:} Divide 125 by 12.
    \item \textit{Output:} 10.4167
\end{itemize}

\noindent
\textbf{Step 3:}
\begin{itemize}
    \item \textit{Subquestion:} Which answer choice matches the output from Step 2?
    \item \textit{Input:} 10.4167, Answer Choices: (A) 2.5498 (B) 0.4167 (C) 3.3987 (D) 8.5497 (E) 5.6312
    \item \textit{Action:} Compare the output with the given answer choices.
    \item \textit{Output:} None of the answer choices match the output from Step 2.
\end{itemize}

It seems there might have been a mistake in the calculation or in the provided answer choices. Let's re-evaluate the calculation in Step 2 to ensure accuracy.

\noindent
\textbf{Step 4:}
\begin{itemize}
    \item \textit{Subquestion:} What is the correct decimal equivalent of 125/12?
    \item \textit{Input:} 125/12
    \item \textit{Action:} Divide 125 by 12 using long division or a calculator.
    \item \textit{Output:} 10.4167 (confirmed)
\end{itemize}

Upon re-evaluation, the calculation is confirmed to be correct, but the output does not match any of the provided answer choices. It's possible that there is an error in the question or the answer choices. However, based on the calculation we've done, none of the answer choices (A) through (E) is correct.

Therefore, among A through E, the answer is none of the provided choices.

\paragraph{COT}: To solve the expression 
\[ \frac{25}{12} \times 5 \], follow these steps:

\textbf{Step 1: Simplify the expression by performing the multiplication.}

\[ \frac{25}{12} \times 5 = \frac{25 \times 5}{12} \]

\textbf{Step 2: Multiply the numerator.}

\[ \frac{25 \times 5}{12} = \frac{125}{12} \]

\textbf{Step 3: Divide 125 by 12 to get the decimal form.}

\[ 125 \div 12 = 10.4167 \]

\textbf{Step 4: Compare the result with the answer choices.}

The result, 10.4167, is not listed among the answer choices. It seems there might have been a mistake in the calculation. Let's try again.

\textbf{Step 1: Simplify the expression by performing the multiplication.}

\[ \frac{25}{12} \times 5 = \frac{25 \times 5}{12} \]

\textbf{Step 2: Multiply the numerator.}

\[ \frac{25 \times 5}{12} = \frac{125}{12} \]

\textbf{Step 3: Divide 125 by 12 to get the decimal form.}

\[ 125 \div 12 = 10 \text{ remainder } 5 \]

To express the remainder as a decimal, we divide 5 by 12:

\[ 5 \div 12 = 0.4167 \]

Adding this to the whole number part (10), we get:

\[ 10 + 0.4167 = 10.4167 \]

This result still does not match any of the answer choices. However, upon re-evaluating the answer choices, it seems that none of them are correct if the original expression was indeed 
\[ \frac{25}{12} \times 5 \]. It's possible that there might be a typo in the question or the answer choices.

Given the calculations we've performed, none of the answer choices (A) through (E) match the correct result of 10.4167. Therefore, it appears there is an error in the provided information.

\section{Errors from Date Understanding}
\subsection{Example 1}
\textit{Question}:2015 is coming in 36 hours. What is the date 24 hours later in MM/DD/YYYY? 

\textit{Answer Choices}: (A) 12/30/2014 (B) 01/01/2015 (C) 01/04/2015 (D) 12/30/2059 (E) 12/31/2014 (F) 01/09/2015 

\textit{GT}:  "A"

\paragraph{IAO}:
\noindent
\textbf{Step 1:}
\begin{itemize}
    \item \textit{Subquestion:} What is the current date if 2015 is coming in 36 hours?
    \item \textit{Input:} 2015 is coming in 36 hours.
    \item \textit{Action:} Subtract 36 hours from the start of 2015 to find the current date.
    \item \textit{Output:} The current date is December 30, 2014.
\end{itemize}

\noindent
\textbf{Step 2:}
\begin{itemize}
    \item \textit{Subquestion:} What is the date 24 hours after the current date?
    \item \textit{Input:} The current date is December 30, 2014.
    \item \textit{Action:} Add 24 hours to the current date to find the date 24 hours later.
    \item \textit{Output:} The date 24 hours later is December 31, 2014.
\end{itemize}

Therefore, among the answer choices A through F, the answer is (E) December 31, 2014.

\paragraph{COT}: If 2015 is coming in 36 hours, we can determine the current date by subtracting 36 hours from the start of 2015.

\begin{enumerate}
\item Since there are 24 hours in a day, 36 hours is equivalent to $1 \text{ day} + 12 \text{ hours}$.
\item The start of 2015 is at $00:00$ on January 1, 2015.
\item Subtracting $1$ day from January 1, 2015, takes us back to December 31, 2014.
\item Since we need to subtract $12$ more hours, and we are currently at $00:00$ on December 31, 2014, we would still be on December 31, 2014, because we are not going back an entire day, just $12$ hours into the previous day.
\end{enumerate}

So, the current date and time, 36 hours before 2015, is $12:00 \text{ (noon)}$ on December 31, 2014.

Now, we need to determine the date 24 hours later:

\begin{enumerate}
\item Adding $24$ hours to $12:00$ on December 31, 2014, will bring us to $12:00$ on January 1, 2015.
\end{enumerate}

Therefore, among A through F, the answer is (B) $01/01/2015$.

\subsection{Example 2}
\textit{Question}:Tomorrow is 11/12/2019. What is the date 10 days ago in MM/DD/YYYY? 

\textit{Answer Choices}: (A) 10/23/2019 (B) 11/02/2019 (C) 10/27/2019 (D) 11/01/2048 (E) 11/01/2019  

\textit{GT}:  "E"

\paragraph{IAO}: \noindent
\textbf{Step 1:}
\begin{itemize}
    \item \textit{Subquestion:} \underline{What is the date 10 days before 11/12/2019?}
    \item \textit{Input:} The given date is 11/12/2019.
    \item \textit{Action:} Subtract 10 days from the given date.
    \item \textit{Output:} The date 10 days before 11/12/2019 is 11/02/2019.
\end{itemize}

\noindent
Therefore, among A through E, the answer is (B) 11/02/2019.

\paragraph{COT}: To find the date 10 days before 11/12/2019, we need to subtract 10 days from that date.

\textbf{Step 1: Start with the given date:} 11/12/2019 (which is in MM/DD/YYYY format, meaning November 12, 2019).

\textbf{Step 2: Subtract 10 days from the 12th of November.} Since $12 - 10 = 2$, if November had more than 2 days, we would end up in November itself. However, since we're subtracting more days than there are in the month up to the 12th, we need to go back to the previous month, which is October.

\textbf{Step 3: October has 31 days.} To find out what date it would be 10 days before November 12, we need to count back from October 31.

\textbf{Step 4: Since we need to go back 10 days from November 12, and there are only 1 day of November included in this count (November 12 itself), we subtract the remaining 9 days from the total number of days in October.}

\textbf{Step 5:} $31 \text{ (total days in October)} - 9 \text{ (days to subtract)} = 22$.

\textbf{Step 6: Therefore, 10 days before November 12, 2019, would be October 22, 2019.}

\textbf{Step 7: Convert this date into MM/DD/YYYY format, which gives us 10/22/2019.}

However, none of the answer choices match this date. It seems there might be a mistake in the answer choices provided or in the calculation. Let's double-check the calculation:

\textbf{Given date:} 11/12/2019

\textbf{Subtract 10 days:}

* November has 12 days up to and including the 12th.
* We need to subtract 10 days.
* $12 - 10 = 2$, so we would be at November 2nd if we didn't need to go into the previous month.
* But since we're subtracting 10 days, we do need to go into the previous month.
* October has 31 days, so the day before November 1st is October 31st.
* We've already accounted for 1 day in November, so we subtract 1 from 10, leaving us with 9 days to subtract from October 31st.
* $31 - 9 = 22$, so we end up on October 22nd.

Upon re-evaluating the calculation, the correct date is indeed October 22, 2019, which is not listed in the answer choices. It seems there may be a typo in the question or the answer choices. However, based on the calculation, the correct answer is not present in the options provided.

\textbf{Therefore, among A through E, the answer is none of the above.}

\section{Errors from StrategyQA}
\subsection{Example 1}
\textit{Question}: Would John Muir not likely have a vitamin D deficiency? 

\textit{GT}:  "yes"

\paragraph{IAO}: To determine whether John Muir would likely have a vitamin D deficiency, we need to consider factors that influence vitamin D levels, such as exposure to sunlight, diet, skin pigmentation, age, geographic location, and lifestyle. Since we don't have specific information about John Muir's personal details, we'll have to rely on general knowledge about his life and activities.

John Muir (1838--1914) was a Scottish-American naturalist, author, environmental philosopher, and early advocate for the preservation of wilderness in the United States. He spent a significant amount of time outdoors, particularly in places like Yosemite and other parts of the American West.

\begin{enumerate}
    \item \textbf{Step 1:}
    \begin{itemize}
        \item \textit{Subquestion:} Did John Muir spend a significant amount of time outdoors?
        \item \textit{Input:} Historical knowledge about John Muir's life and activities.
        \item \textit{Action:} Assess the likelihood of outdoor activity based on known biographical information.
        \item \textit{Output:} Yes, John Muir spent a significant amount of time outdoors.
    \end{itemize}

    \item \textbf{Step 2:}
    \begin{itemize}
        \item \textit{Subquestion:} Does spending a significant amount of time outdoors increase exposure to sunlight?
        \item \textit{Input:} General knowledge about outdoor activities and sunlight exposure.
        \item \textit{Action:} Correlate outdoor activity with sunlight exposure.
        \item \textit{Output:} Yes, spending a significant amount of time outdoors generally increases exposure to sunlight.
    \end{itemize}

    \item \textbf{Step 3:}
    \begin{itemize}
        \item \textit{Subquestion:} Does increased exposure to sunlight affect vitamin D levels?
        \item \textit{Input:} Scientific knowledge about vitamin D synthesis.
        \item \textit{Action:} Relate sunlight exposure to vitamin D synthesis.
        \item \textit{Output:} Yes, increased exposure to sunlight typically leads to higher vitamin D levels due to the skin's production of vitamin D in response to UVB rays.
    \end{itemize}

    \item \textbf{Step 4:}
    \begin{itemize}
        \item \textit{Subquestion:} Based on John Muir's outdoor activities and the relationship between sunlight exposure and vitamin D synthesis, is it likely that John Muir had a vitamin D deficiency?
        \item \textit{Input:} Outputs from Step 1, Step 2, and Step 3.
        \item \textit{Action:} Infer the likelihood of vitamin D deficiency based on the correlation between outdoor activity, sunlight exposure, and vitamin D synthesis.
        \item \textit{Output:} It is unlikely that John Muir had a vitamin D deficiency given his significant time spent outdoors and the resulting increased exposure to sunlight.
    \end{itemize}
\end{enumerate}

Therefore, the answer (Yes or No) is No.

\paragraph{COT}: \textbf{Step 1: Identify the factors that contribute to vitamin D deficiency.}

Vitamin D deficiency can be caused by several factors, including:

\begin{itemize}
\item Lack of exposure to sunlight
\item Inadequate dietary intake of vitamin D
\item Certain medical conditions
\item The body's inability to absorb or convert vitamin D to its active form
\end{itemize}

\textbf{Step 2: Consider John Muir's lifestyle.}

John Muir was a naturalist, explorer, and advocate for the preservation of wilderness.  His lifestyle involved spending a significant amount of time outdoors, often in sunny environments like California's Sierra Nevada mountains.

\textbf{Step 3: Assess the likelihood of vitamin D deficiency based on lifestyle.}

Given that John Muir spent a lot of time outdoors, he would have been exposed to sunlight regularly. Sunlight exposure is a primary source of vitamin D, as the body synthesizes vitamin D when the skin is exposed to UV rays from the sun.

\textbf{Step 4: Draw a conclusion based on the information.}

Since John Muir was frequently outdoors and exposed to sunlight, it is \emph{less likely} that he would have suffered from vitamin D deficiency due to lack of sunlight exposure.

\textbf{Therefore, the answer (Yes or No) is No.}


\textit{Question}: Can French Defence initial move defend against four move checkmate? 

\textit{GT}:  "no"

\paragraph{IAO}: To determine if the French Defence can defend against the four-move checkmate (also known as the Scholar's Mate), we need to analyze the moves involved in both the French Defence and the Scholar's Mate.

\begin{itemize}
    \item \textbf{Step 1:}
    \begin{itemize}
        \item \textit{Subquestion:} What are the typical opening moves for the French Defence?
        \item \textit{Input:} The French Defence is a chess opening characterized by the moves.
        \item \textit{Action:} Describe the typical opening moves for the French Defence.
        \item \textit{Output:} The French Defence typically begins with 1.e4 e6.
    \end{itemize}

    \item \textbf{Step 2:}
    \begin{itemize}
        \item \textit{Subquestion:} What are the moves involved in the four-move checkmate?
        \item \textit{Input:} The four-move checkmate (Scholar's Mate) sequence.
        \item \textit{Action:} Describe the typical sequence of moves that lead to the four-move checkmate.
        \item \textit{Output:} The four-move checkmate typically involves the moves 1.e4 e5 2.Qh5 Nc6 3.Bc4 Nf6 4.Qxf7\#.
    \end{itemize}

    \item \textbf{Step 3:}
    \begin{itemize}
        \item \textit{Subquestion:} Can the opening move of the French Defence prevent the specific sequence of the four-move checkmate?
        \item \textit{Input:} The opening move of the French Defence (1.e4 e6) and the sequence of the four-move checkmate (1.e4 e5 2.Qh3 Nc6 3.Bc4 Nf6 4.Qxf7\#).
        \item \textit{Action:} Compare the opening moves of the French Defence with the moves required for the four-move checkmate to determine if the French Defence can prevent the checkmate sequence.
        \item \textit{Output:} The French Defence does not play 1...e5, which is a necessary move for the four-move checkmate. Instead, it plays 1...e6, which already deviates from the Scholar's Mate sequence.
    \end{itemize}

    \item \textbf{Step 4:}
    \begin{itemize}
        \item \textit{Subquestion:} Does the deviation from the four-move checkmate sequence inherently prevent the checkmate from occurring?
        \item \textit{Input:} The deviation in the opening move of the French Defence from the Scholar's Mate sequence.
        \item \textit{Action:} Analyze the impact of the deviation on the possibility of the four-move checkmate occurring.
        \item \textit{Output:} Since the French Defence does not follow the move 1...e5, it inherently prevents the specific sequence of the four-move checkmate from occurring.
    \end{itemize}

\end{itemize}

Therefore, the answer (Yes or No) is Yes.

\paragraph{COT}: The Four-Move Checkmate, also known as the Scholar's Mate, is a quick checkmate that White can deliver after the following moves:

\begin{enumerate}
\item e4 e5
\item Qh5 Nc6
\item Bc4 Nf6
\item Qxf7\#
\end{enumerate}

In this sequence, White aims to attack the f7 square, which is a weak point in Black's position because it is only defended by the king at the beginning of the game.

The French Defense, on the other hand, is a chess opening that begins with the moves:

\begin{enumerate}
\item e4 e6
\end{enumerate}

The idea behind the French Defense is to solidify Black's position and create a strong pawn chain that can later be used to counterattack White's center. The French Defense does not immediately contest the e5 square, which is a key component of setting up the Scholar's Mate.

Let's see if the French Defense can prevent the Four-Move Checkmate:

\begin{enumerate}
\item e4 e6
\item d4 d5
\end{enumerate}

With the second move d5, Black challenges White's center and opens lines for the bishop and queen, which can help in defending against early threats.

If White tries to set up the Scholar's Mate anyway:

\begin{enumerate}
\item 3. Qh5
\end{enumerate}

Black can now defend easily because the move ...e6 has already prevented the bishop from coming to c4 to attack the f7 square. Black has several reasonable moves here, such as ...Nf6, which attacks the invading queen and prepares to castle, or even ...g6, which would force the queen to move again.

Since the French Defense does not allow White to establish the same attacking pattern as in the Scholar's Mate, it effectively prevents the quick checkmate.

\textbf{Therefore, the answer (Yes or No) is Yes.}

\end{document}

%% file: paper/intro.tex
The remarkable capabilities of large language models (LLMs) \cite{vaswani_attention_2017,devlin_bert_2019,raffel2020exploring,brown2020language,chowdhery2023palm} have raised fundamental questions about how these models acquire, represent, and utilize knowledge \cite{ju-etal-2024-large,meng2022locating,turpin2024language,zhang2024knowledge}. While LLMs demonstrate impressive few-shot and zero-shot learning abilities, understanding their knowledge application process remains challenging. \textit{Prompting} \cite{liu2023pre} has emerged as a crucial technique for accessing and directing LLMs' knowledge, leading to extensive research in both manual \cite{schick-schutze-2021-just,reynolds2021prompt} and automated prompting approaches \cite{gao-etal-2021-making,shin-etal-2020-autoprompt}.

\begin{figure}[ht]
\centering
\includegraphics[width=0.7\columnwidth]{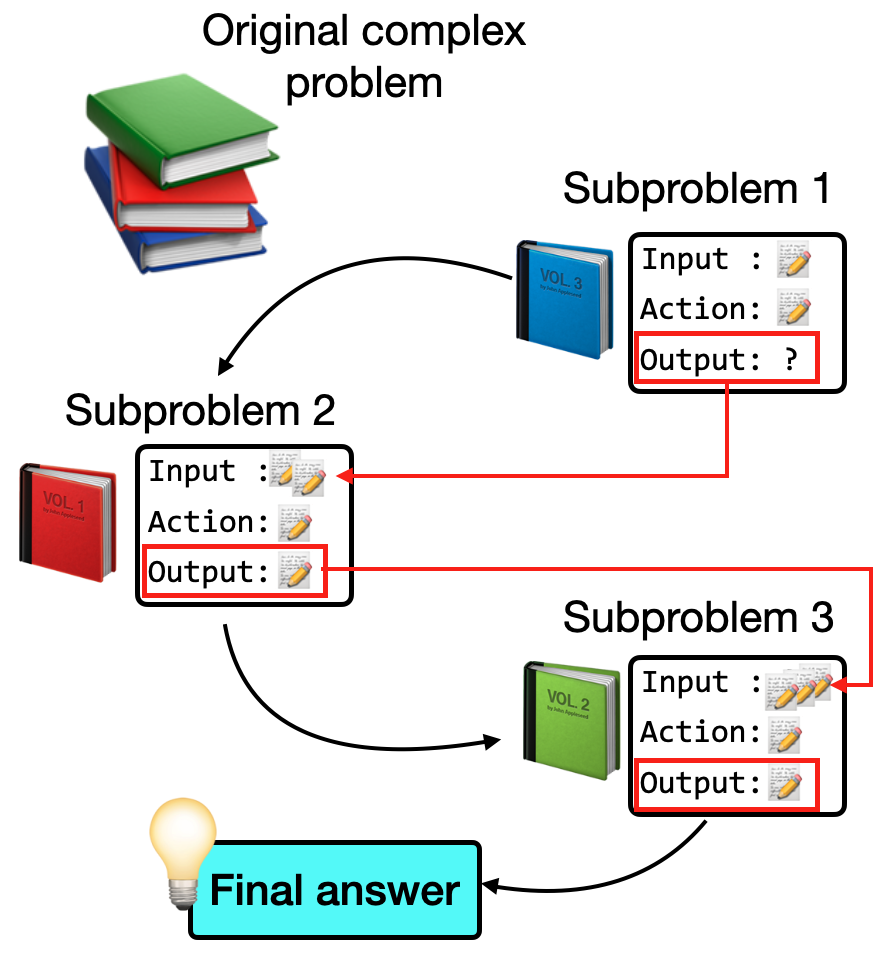}
\label{fig:iao_front}
\caption{Illustration of IAO prompting demonstrating how knowledge is structured and applied through explicit Input-Action-Output steps. Each step's output becomes verified knowledge for subsequent reasoning.}
\end{figure}

Chain-of-Thought (CoT) prompting \cite{wei2022chain,wang-etal-2022-iteratively} represents a significant advancement in accessing LLMs' knowledge by making intermediate reasoning steps explicit. This approach guides LLMs to decompose complex knowledge application into sequential steps, similar to human reasoning processes. In zero-shot settings, simple prompts like "\texttt{let's think step by step}" have proven effective in activating LLMs' inherent reasoning capabilities \citep{kojima2022large}, suggesting that these models possess substantial implicit knowledge that can be systematically accessed.

However, a critical challenge persists: ensuring that LLMs' knowledge utilization is both interpretable and verifiable. As noted by \citet{singh2024rethinking}, understanding an LLM's knowledge application requires extracting and validating the relationships learned by the model. While CoT improves reasoning performance, it often fails to provide a clear mapping of how stored knowledge is accessed and applied in generating outputs step by step.

\begin{figure*}[ht]
\includegraphics[width=\textwidth]{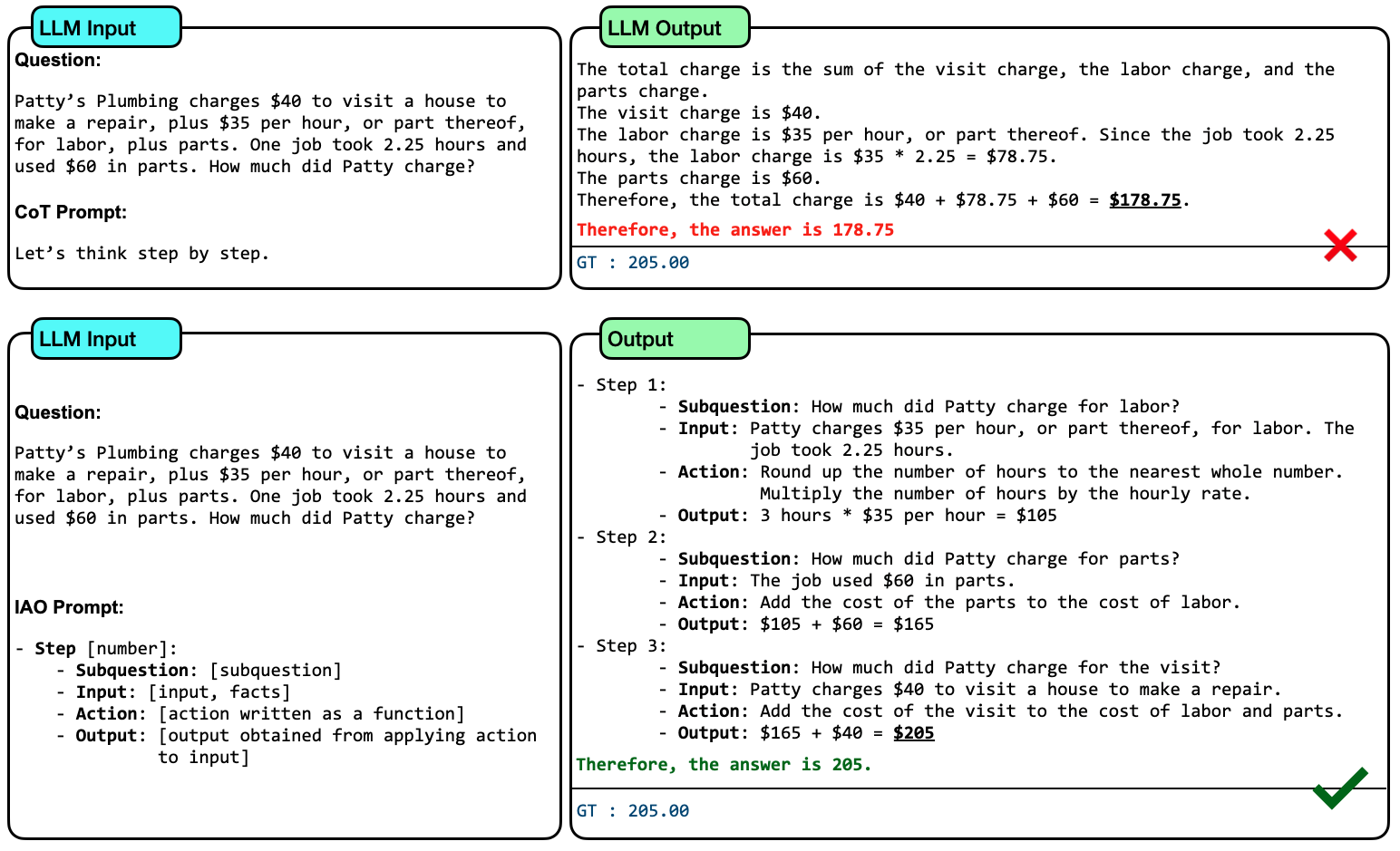}
\caption{Comparison of knowledge application between IAO prompting and zero-shot CoT using PALM-2 on GSM8k. IAO's structured format reveals how knowledge is accessed and applied at each step, while CoT misses crucial information.}
\label{fig:iao_example_fig}
\end{figure*}

This opacity in knowledge utilization poses significant challenges for verifying factual accuracy, identifying knowledge gaps, and ensuring reliable reasoning \cite{chen2024survey}. To address these challenges, we introduce IAO (Input-Action-Output) prompting, a structured template that explicitly models how LLMs access and apply their knowledge. Each reasoning step clearly delineates the input knowledge being used, the knowledge-based action being performed, and the resulting output, creating a transparent chain of knowledge utilization.

Our main contributions are:
(i) we propose IAO prompting, a template-based approach that makes explicit how LLMs access and apply their stored knowledge during reasoning;
(ii) we demonstrate through experiments across various reasoning tasks how IAO enhances knowledge verification through intermediate steps; (iii) through systematic human evaluation, we show that IAO's structured format significantly improves evaluators' ability to identify knowledge gaps, verify reasoning steps, and detect potential hallucinations compared to traditional prompting methods.

%% file: paper/related_work.tex
\paragraph{Prompting}
Chain-of-thought (CoT) prompting represents a fundamental approach to accessing and utilizing LLMs' stored knowledge through intermediate reasoning steps. While zero-shot CoT \cite{kojima2022large} demonstrates LLMs' ability to access implicit knowledge without examples, few-shot CoT \cite{wei2022chain} requires explicit knowledge demonstrations. \citet{zhang2022automatic} proposed automating the construction of knowledge demonstrations, reducing manual effort while maintaining knowledge fidelity. Recent work has explored structuring knowledge representation in prompting through programs \cite{chen2023program}, knowledge graphs \cite{besta2024graph}, and tabular formats \cite{ziqi-lu-2023-tab}. Our work focuses on making knowledge utilization explicit in free-text reasoning chains, providing a transparent view of how LLMs access and apply their stored knowledge.

\paragraph{Problem Decomposition in LLMs}

Decomposing complex knowledge tasks into smaller units enhances LLMs' knowledge utilization capabilities \cite{shridhar2022automatic}. The Least-to-Most (L2M) prompting approach \cite{zhou2022least} improves reasoning by sequentially solving simpler sub-problems incrementally from least to most complex. Similarly, the Plan-and-Solve framework \cite{wang2023plan} first generates a high-level plan outlining the solution path before solving the problem step by step, enhancing zero-shot chain-of-thought reasoning. Tab-CoT \cite{ziqi-lu-2023-tab} organizes knowledge application steps in a tabular format to enhance visibility of knowledge flow but often requires domain-specific engineering. IAO prompting differs by autonomously decomposing knowledge application while explicitly tracking knowledge flow through input-action-output sequences, enabling better verification of knowledge utilization at each step.

\paragraph{Knowledge Verification in LLMs}
Ensuring factual accuracy and verifying knowledge application in LLMs remains a critical challenge. Early research focused on specific domains, such as dialogue systems \cite{shuster2021retrieval} and question-answering tasks \cite{kadavath2022language,kandpal2023large}. Recent work has expanded to broader verification challenges, particularly addressing the challenge of long-tail knowledge \cite{kandpal2023large} and developing methods to assess when LLM outputs can be trusted \cite{mallen2022not}. Of particular relevance to our work is the emergence of self-verification approaches, exemplified by \cite{manakul2023selfcheckgpt}, which enable zero-shot detection of hallucinations without external knowledge sources. Our IAO approach complements these verification efforts by providing a structured framework that makes knowledge application steps explicit and independently verifiable, helping to identify potential knowledge gaps or misapplications at each reasoning step. Unlike previous works that focus on post-hoc verification or external knowledge sources, IAO enforces verification during the reasoning process itself through its structured input-action-output format, making the knowledge flow transparent and checkable in real-time as the model generates its response.

%% file: paper/approach.tex
We introduce IAO, a prompting technique that explicitly structures how LLMs access, process, and apply their stored knowledge through a decomposed reasoning chain. This approach is inspired by knowledge-based procedural reasoning, where complex problems are solved by systematically accessing and applying relevant knowledge components. Unlike implicit knowledge utilization, IAO enforces an explicit use of knowledge flow through each reasoning step.

The motivation stems from observing that zero-shot CoT often produces incomplete knowledge chains or incorrectly merges knowledge application steps, leading to errors. As illustrated in \Cref{fig:iao_example_fig}, 
traditional approaches may overlook crucial knowledge elements (such as \textit{or part thereof}) by failing to systematically process available information. IAO's structured framework improves both knowledge utilization and verification by making explicit which knowledge components are being accessed and applied at each step.

IAO prompting comprises four knowledge-focused components as shown in \Cref{fig:iao_example_fig}:

\begin{enumerate}
    \item \textbf{Subquestion}: The LLM breaks down the primary question into smaller knowledge components, creating a sequence of targeted knowledge queries that build upon each other.
    
    \item \textbf{Input}: For each step, the LLM explicitly identifies the relevant knowledge components available, including both given facts and previously derived knowledge.
    
    \item \textbf{Action}: The LLM specifies how it plans to apply or transform the input knowledge, often expressing this as a knowledge operation or procedure.
    
    \item \textbf{Output}: The LLM generates new knowledge by applying the planned operation to the input knowledge. This derived knowledge becomes available for subsequent reasoning steps.
\end{enumerate}

This structured approach enables better tracking of knowledge flow, verification of knowledge application, and identification of potential knowledge gaps or misapplications. In the following sections, we empirically evaluate how this knowledge-structured approach enhances reasoning across various tasks.

%% file: paper/experiments.tex
\begin{table*}[!t]
\resizebox{\textwidth}{!}{%
\begin{tabular}{@{}lccccccc@{}}
\toprule
\multirow{2}{*}{Knowledge Domain} &
  \multicolumn{2}{c}{\textit{Arithmetic Knowledge}} &
  \multicolumn{2}{c}{\textit{Logical Knowledge}} &
  \multicolumn{2}{c}{\textit{Commonsense Knowledge}} &
  \textit{Symbolic Knowledge} \\ \cmidrule(l){2-8} 
 &
  \multicolumn{1}{l}{AQUA} &
  \multicolumn{1}{l}{GSM8K} &
  Date Understanding &
  Object Tracking &
  \multicolumn{1}{l}{StrategyQA} &
  \multicolumn{1}{l}{CommonsenseQA} &
  Last Letter \\ \midrule
zero-shot CoT        & 61.8 & 76.4 & 85.3 & 61.2 & 72.8 & 78.4 & 75.6 \\
zero-shot IAO (ours) & 63.9 & \textbf{82.3} & \textbf{88.1} & {67.1} & \textbf{76.9} & \textbf{83.1} & \textbf{88.8} \\ \midrule
L2M              & \textbf{64.2} & 77.5 & 86.4 & 63.8 & 73.5 & 79.2 & 83.2 \\
P\&S             & 62.7 & 79.3 & 87.2 & \textbf{67.5} & 74.8 & 80.6 & 84.5 \\  \bottomrule

\end{tabular}%
}
\caption{Evaluation results for PALM-2. Bold denotes best result. All methods use the same answer extraction prompt in a single stage for fair comparison. All methods are evaluated under the zero-shot setting.}

\label{tab:palm_results}
\end{table*}

\begin{table*}[!t]
\resizebox{\textwidth}{!}{%
\begin{tabular}{@{}lccccccc@{}}
\toprule
\multicolumn{1}{c}{\multirow{2}{*}{Knowledge Domain}} &
  \multicolumn{2}{c}{\textit{Arithmetic Knowledge}} &
  \multicolumn{2}{c}{\textit{Logical Knowledge}} &
  \multicolumn{2}{c}{\textit{Commonsense Knowledge}} &
  \textit{Symbolic Knowledge} \\ \cmidrule(l){2-8} 
\multicolumn{1}{c}{} & AQUA          & GSM8K         & Date Understanding & Object Tracking & StrategyQA     & CommonsenseQA & Last Letter   \\ \midrule
zero-shot CoT        & 68.4 & 90.1 & 81.5 & 97.8 & 73.2 & 79.5 & 90.3 \\
zero-shot IAO (ours) & 70.2 & \textbf{94.2} & {83.2} & {100} & \textbf{76.3} & \textbf{84.8} & \textbf{94.7} \\ \midrule
L2M              & 69.1 & 91.8 & 82.4 & 100 & 74.8 & 81.2 & 92.6 \\
P\&S             & \textbf{73.1} & 92.4 & \textbf{83.8} & 100 & 75.1 & 83.7 & 93.8 \\
\bottomrule
\end{tabular}%
}
\caption{Evaluation results for GPT-4. Bold denotes best result. All methods use the same answer extraction prompt in a single stage for fair comparison. All methods are evaluated under the zero-shot setting.}

\label{tab:gpt4_results}
\end{table*}

\paragraph{Tasks}

To evaluate IAO, we use common datasets across four different reasoning tasks: (a) \textit{Mathematical reasoning} (using GSM8k \cite{cobbe2021training} and AQuA \cite{ling2017program}) for testing numerical computation and mathematical reasoning capabilities; 
(b) \textit{Commonsense reasoning} (using StrategyQA \cite{geva2021did} and  CommonsenseQA \cite{talmor2019commonsenseqa}); (c) \textit{Symbolic reasoning} (using Last Letter \cite{wei2022chain}) for  evaluating abstract pattern recognition and manipulation; (d) \textit{Logical Reasoning} using Date Understanding and Shuffled Object Tracking \cite{srivastava2023beyond}) for testing temporal reasoning and spatial relationship understanding.

Dataset statistics are presented in \Cref{tab:data_statistics}, with additional details in \Cref{tab:more_data_stats}.

\paragraph{Models}

We conduct the experiments using (a) PALM-2 \cite{anil2023palm} (\texttt{text-unicorn}) which provides insights into knowledge utilization in mid-size models and GPT-4 (\texttt{gpt-4-1106-preview}) \cite{achiam2023gpt} that represents advanced knowledge capabilities.

Initial experiments with GPT-3.5 revealed limitations in following structured knowledge templates, highlighting how model scale affects knowledge organization capabilities. Our template-based approach demonstrates the ability to guide knowledge application without task-specific demonstrations, suggesting effective knowledge structuring across different model scales.

\paragraph{Baselines}

As a baseline, we compare our approach to chain-of-thought (CoT) \cite{wei2022chain}, in particular zero-shot-CoT \cite{kojima2022large} with the prompt \texttt{Let's think step by step} appended to the question. We use two different settings: the single step where the reasoning prompt and answer extraction prompt are in a single API call; and the two-step setting.

Additionally, we compare IAO to relevant prompting frameworks focusing on problem decomposition and planning before solving the problem to substantiate IAO's validity. In particular, we compare to \cite{zhou2022least} as \textbf{L2M} and \cite{wang2023plan} as \textbf{P\&S}.  
It is worth mentioning that these methods are multi-prompting approaches which is not the case for IAO.

%% file: paper/results.tex

\subsection{Arithmetic Reasoning} 

\paragraph{GPT-4} As shown in \Cref{tab:gpt4_results}, IAO demonstrates notable improvements in arithmetic reasoning tasks, particularly on the GSM8K dataset where it achieves 94.2\% accuracy, outperforming the zero-shot CoT baseline (90.1\%) by 4.1 percentage points. This significant improvement suggests that IAO's structured decomposition particularly benefits complex multi-step arithmetic problems, which are characteristic of GSM8K. For the AQUA dataset, IAO achieves 70.2\% accuracy, showing moderate improvement over the zero-shot CoT baseline (68.4\%). While P\&S achieves the highest performance on AQUA (73.1\%), IAO consistently outperforms both zero-shot CoT and L2M baselines across both arithmetic datasets, demonstrating the effectiveness of explicit input-action-output structuring in mathematical reasoning tasks.
 Moreover, we observe that most cases where the IAO prompt fails are due to inherent arithmetic calculation failures. Loosely speaking, while the ``action'' planned is correct and the function is also correct, the LLM does calculation errors or fails to report the correct output as a final answer. We discuss this further in the following sections. 

\paragraph{PALM-2} We observe the same trend when using PALM-2.
There is a decrease in terms of percentage points (p.p), 2.1\%, for the AQuA dataset (where the zero-shot CoT baseline accuracy is of 61.8\%) but improves over the zero-shot CoT baseline for the GSM8k dataset (5.9 p.p). The same observations about the errors and failures for the IAO prompting apply to PALM-2 too.

\subsection{Logical Reasoning}
The datasets studied are Date Understanding and Object tracking from \cite{srivastava2023beyond}. The former asks the models
to infer the date from a context. Tracking Shuffled Objects tests a model’s ability to infer the final state of objects given its initial state and a sequence of object shuffling. 
\paragraph{GPT-4} GPT-4 achieved perfect scores for Object tracking task for almost all baselines settings with the exception of zero-shot CoT. However, that is not the case for the Date Understanding task where the baseline accuracy for zero-shot CoT is of 81.5 while IAO achieves 83.2 p.p. The challenge in this task is to correctly interpret the question and understand the temporal setting. Even by forcing the LLM to interrogate itself about what it knows (\textit{Input} field) it fails to solve this issue.  
\paragraph{PALM-2} IAO prompting achieves 67.1 p.p (over a zero-shot CoT baseline of 61.2) for the Object Tracking dataset. On the other hand, it achieves and accuracy of 88.1 from 85.3 for the same baseline prompt for the Date Understanding task. 

\subsection{Commonsense Reasoning}
CommonsenseQA asks questions with complex semantics that often require reasoning based on prior knowledge \cite{talmor2019commonsenseqa}. StrategyQA  dataset have implicit multi-step questions requiring a wide-range
of reasoning skills to answer them \cite{geva2021did}.
\paragraph{GPT-4} For the StrategyQA dataset, IAO improves the accuracy metric over all baselines but by modest margin. This is likely because the dataset contains a number of ambiguous questions, which if read verbatim may have many plausible answers but the ground truth contains only one answer. GPT-4 will in such cases refrain from giving a final answer, which is counted as an incorrect answer. This lead to a task accuracy of 76.3 p.p in StrategyQA and 84.8 in CommonsenseQA where the zero-shot CoT prompt accuracy are 73.2 and 79.5 respectively.

\paragraph{PALM-2} For StrategyQA, we see an improvement from 72.8 p.p to 76.9. For CommonsenseQA, we observe an increase in accuracy over the the zero-shot CoT baseline of 4.7 p.p.

\subsection{Symbolic Reasoning} We use the Last Letter Concatenation \cite{wei2022chain} dataset which contains questions asking the model to concatenate the last letters of each word. 
\paragraph{GPT-4} GPT-4 reaches 90.3 p.p in the zero-shot CoT setting. IAO prompting improves over all the baselines reaching 94.7 p.p in accuracy.  
\paragraph{PALM-2} Here the gains are more substantial. IAO prompting improves the accuracy by 11.6 p.p reaching 88.8 p.p over a baseline of 75.6 p.p. From the examples we observe how dividing the problem into multiple subproblems and solving each one in a structured way increases the model's reasoning abilities.

\paragraph{Comparison to L2M and P\&S} As shown in row 3 and 4 \Cref{tab:gpt4_results} and \Cref{tab:palm_results} IAO shows improvement over these baselines approaches on both GPT-4 and PaLM-2 across almost all the evaluated tasks with the additional benefit of added transparency and interpretability of the reasoning steps in a single-step prompting approach.

\subsection{Ablation studies}

We additionally perform ablation studies by removing parts of the template and assessing performance on specific tasks. Through this process, we aim to achieve three key objectives: (i) isolate the impact of each field in the proposed template, (ii) identify redundancies and (iii) enhance interpretability. The results are presented in \Cref{tab:ablation_results} and \Cref{tab:ablation_results_appendix}.

\begin{table}[h]
\resizebox{\columnwidth}{!}{%
\begin{tabular}{@{}lc@{}}
\toprule
Prompt                                              & \multicolumn{1}{c}{Average} \\ \midrule
\texttt{[Step, Input, Action, Output]}              & 68.9                        \\
\texttt{[Step, Subquestion, Action, Output]}         & 77.7                        \\
\texttt{[Step, Subquestion, Input, Output]}         & 76.3                        \\
\texttt{[Step, Subquestion, Input, Action]}         & 59.0                        \\ \midrule
\texttt{[Step, Subquestion, Input, Action, Output]} & 80.9                        \\ \bottomrule
\end{tabular}%
}
\caption{Performance in a zero-shot setting when one of the field of the prompt is removed.}
\label{tab:ablation_results}
\end{table}

First, we observe that the prompt with all fields (\texttt{[Step, Subquestion, Input, Action, Output]}) achieves the highest average performance (80.9 p.p), indicating that including all available information is beneficial. Including the \texttt{Subquestion} field consistently improves performance compared to excluding it, suggesting that it is important for the model to understand the context of the task. Removing the \texttt{Output} from the prompt leads to a significant drop in performance (12.0 p.p). This suggests that the model is able to use the output information from previous steps to improve its performance on subsequent steps. Removing the \texttt{Action} also results in a noticeable decrease in performance (3.2 p.p). This suggests that the model is able to use the action information to better understand the context of the task.

%% file: paper/discussion.tex
\begin{table*}[!t]
\resizebox{\textwidth}{!}{%
\begin{tabular}{@{}llllll@{}}
\toprule
\multicolumn{2}{l}{Question} &
  \multicolumn{4}{l}{\begin{tabular}[c]{@{}l@{}} \textit{The original price of an item is discounted 22\%. A customer buys the item at this discounted price using a \$20-off coupon. There is no tax on the item, and this} \\ \textit{was the only item the customer bought. If the customer paid \$1.90 more than half the original price of the item, what was the original price of the item?} \\ Answer Choices: (A) \$61 (B) \$65 (C) \$67.40 (D) \$70 (E) \$78.20 \hspace{5cm}\textcolor{blue}{(GT: (E))}\end{tabular}} \\ \midrule \midrule
\multirow{6}{*}{\rotatebox[origin=c]{90}{PALM-2}} &
  CoT &
  \multicolumn{4}{l}{\begin{tabular}[c]{@{}l@{}}Let x be the original price of the item. The discounted price is 0.78x. The customer paid 0.78x - 20. This is \$1.90 more than half the original price, which is 0.5x + 1.90. \\ So, 0.78x - 20 = 0.5x + 1.90. Solving for x, we get x = 70.\\ Therefore, among A through E, the answer is \textbf{(D)}.\end{tabular}} \\ \cmidrule(l){2-6} 
 &
  \multirow{5}{*}{IAO} &
  Subquestion &
  Input &
  Action &
  Output \\ \cmidrule(l){3-6} 
 &
   &
  1. What is the discounted price of the item? &
  Original price = x, Discount = 22\% &
  \begin{tabular}[c]{@{}l@{}}Discounted price = \\ Original price - (Original price * Discount)\end{tabular} &
  \begin{tabular}[c]{@{}l@{}}Discounted price =\\  x - (x * 0.22) = 0.78x\end{tabular} \\
 &
   &
  2. What is the price the customer paid for the item? &
  Discounted price = 0.78x, Coupon = \$20 &
  Price paid = Discounted price - Coupon &
  Price paid = 0.78x - 20 \\
 &
   &
  \begin{tabular}[c]{@{}l@{}}3. What is the equation that related the price \\ paid to the original price?\end{tabular} &
  \begin{tabular}[c]{@{}l@{}}3. Price paid = 0.78x - 20, \\ Price paid = 0.5x + 1.90\end{tabular} &
  \begin{tabular}[c]{@{}l@{}}Set the two expressions for the price \\ paid equal to each other.\end{tabular} &
  0.78x - 20 = 0.5x + 1.90 \\
 &
   &
  4. What is the original price of the item? &
  0.78x - 20 = 0.5x + 1.90 &
  Solve for x. &
  \textbf{x = 78.20} \\ \midrule \midrule 
\multirow{7}{*}{\rotatebox[origin=c]{90}{GPT-4}} &
  CoT &
  \multicolumn{4}{l}{\begin{tabular}[c]{@{}l@{}}Let's denote the original price of the item as \( P \).\\ The item is discounted by 22\%, so the customer pays 78\% of the original price. We can express this as \( 0.78P \).\\ Then, the customer uses a \$20-off coupon, so we subtract \$20 from the discounted price: \( 0.78P - 20 \).\\ According to the problem, the customer paid \$1.90 more than half the original price of the item.\\ Half the original price is \( \frac{P}{2} \), so the amount the customer paid can also be expressed as \( \frac{P}{2} + 1.90 \).\\ Now we can set up the equation: \( 0.78P - 20 = \frac{P}{2} + 1.90 \)\\ To solve for \( P \), we first get rid of the fractions by multiplying every term by 2: \( 2(0.78P) - 2(20) = 2(\frac{P}{2}) + 2(1.90) \)\\ \( 1.56P - 40 = P + 3.80 \)\\ Now, we subtract \( P \) from both sides to get: \( 0.56P - 40 = 3.80 \)\\ Next, we add 40 to both sides: \( 0.56P = 43.80 \)\\ Finally, we divide both sides by 0.56 to solve for \( P \):\\ \( P = \frac{43.80}{0.56} \)\\ \( P = 78.2142857... \)\\ The closest answer choice to \$78.21 is \$78.20.\\ Therefore, among A through E, the answer is \textbf{(E) \$78.20)} .\end{tabular}} \\ \cmidrule(l){2-6} 
 &
  \multirow{6}{*}{IAO} &
  Subquestion &
  Input &
  Action &
  Output \\ \cmidrule(l){3-6} 
 &
   &
  \begin{tabular}[c]{@{}l@{}}1. What is the equation that represents \\ the relationship between the original price, \\ the discount, the coupon, and the amount \\ the customer paid?\end{tabular} &
  \begin{tabular}[c]{@{}l@{}}Let the original price be P. The customer \\ gets a 22\% discount and then uses a \\ \$20-off coupon. The customer paid\\ \$1.90 more than half the original price.\end{tabular} &
  \begin{tabular}[c]{@{}l@{}}Write the equation based on the given\\  information.\end{tabular} &
  P - 0.22P - 20 = 0.5P + 1.90 \\
 &
   &
  \begin{tabular}[c]{@{}l@{}}2. How can we simplify the equation from S\\ tep 1 to solve for P?\end{tabular} &
  P - 0.22P - 20 = 0.5P + 1.90 &
  Combine like terms and isolate P. &
  0.78P - 20 = 0.5P + 1.90 \\
 &
   &
  3. What is the value of P? &
  0.78P - 20 = 0.5P + 1.90 &
  Solve for P. &
  P = (20 + 1.90) / (0.78 - 0.5) \\
 &
   &
  4. What is the numerical value of P? &
  (20 + 1.90) / (0.78 - 0.5) &
  Perform the arithmetic to find P. &
  P = 21.90 / 0.28 \\
 &
   &
  \begin{tabular}[c]{@{}l@{}}5. What is the final calculation to determine \\ the original price?\end{tabular} &
  21.90 / 0.28 &
  Calculate the value of P. &
  \textbf{P = 78.2142857...} \\ \bottomrule
\end{tabular}%
}
\caption{Output example from AQuA dataset. The final results are in \textbf{bold}.}
\label{tab:aqua_example}
\end{table*}

\begin{table*}[]
\resizebox{\textwidth}{!}{%
\begin{tabular}{@{}llllll@{}}
\toprule
\multicolumn{2}{l}{Question} &
  \multicolumn{4}{l}{\textit{Could all of the people who pass through 30th Street Station every day fit in Dorton Arena?}                               \hspace{5cm} \textcolor{blue}{(GT: (No))}·} \\ \midrule \midrule
\multirow{5}{*}{\rotatebox[origin=c]{90}{PALM-2}} &
  CoT &
  \multicolumn{4}{l}{\begin{tabular}[c]{@{}l@{}}30th Street Station is the busiest train station in Philadelphia, with an average of 120,000 passengers passing through each day. Dorton Arena is a multi-purpose arena in \\ Raleigh, North Carolina, with a seating capacity of 5,000 people.\\ Therefore, the answer is No\end{tabular}} \\ \cmidrule(l){2-6} 
 &
  \multirow{4}{*}{IAO} &
  Subquestion &
  Input &
  Action &
  Output \\ \cmidrule(l){3-6} 
 &
   &
  \begin{tabular}[c]{@{}l@{}}1. How many people pass through \\ 30th Street Station every day?\end{tabular} &
  \begin{tabular}[c]{@{}l@{}}30th Street Station is the busiest train \\ station in Philadelphia, with an average of \\ 120,000 passengers passing through each day.\end{tabular} &
  \begin{tabular}[c]{@{}l@{}}The number of people who pass \\ through 30th Street Station \\ every day is 120,000.\end{tabular} &
  120,000 \\
 &
   &
  2. What is the capacity of Dorton Arena? &
  \begin{tabular}[c]{@{}l@{}}Dorton Arena is a multi-purpose arena located \\ in Raleigh, North Carolina. It has a seating \\ capacity of 10,000 people.\end{tabular} &
  \begin{tabular}[c]{@{}l@{}}The capacity of Dorton Arena is \\ 10,000 people.\end{tabular} &
  10,000 \\
 &
   &
  \begin{tabular}[c]{@{}l@{}}3. Can all of the people who pass through \\ 30th Street Station every day fit in \\ Dorton Arena?\end{tabular} &
  \begin{tabular}[c]{@{}l@{}}The number of people who pass through 30th\\ Street Station every day is 120,000. The \\ capacity of Dorton Arena is 10,000 people.\end{tabular} &
  \begin{tabular}[c]{@{}l@{}}The number of people who pass \\ through 30th Street Station every day \\ is greater than the capacity of Dorton Arena.\end{tabular} &
  \textbf{No} \\ \midrule \midrule
\multirow{7}{*}{\rotatebox[origin=c]{90}{GPT-4}} &
  \multirow{2}{*}{CoT} &
  \multicolumn{4}{l}{\begin{tabular}[c]{@{}l@{}}To determine if all of the people who pass through 30th Street Station every day could fit in Dorton Arena, we need to know two key pieces of information:\\ 1. The average daily number of people passing through 30th Street Station.\\ 2. The seating capacity of Dorton Arena.\\ Once we have these two figures, we can compare them to see if the arena can accommodate the daily passengers of 30th Street Station. \\ Let's assume we have the necessary data:\\ - The average daily number of people passing through 30th Street Station (we'll call this number X).\\ - The seating capacity of Dorton Arena (we'll call this number Y).\\ If X is less than or equal to Y, then the answer is Yes, all of the people can fit in Dorton Arena. If X is greater than Y, then the answer is No, they cannot all fit. \\ Since I don't have the current data, you would need to provide the average daily passenger count for 30th Street Station and the seating capacity of Dorton Arena \\ for me to complete this comparison.\end{tabular}} \\ \cmidrule(l){3-6} 
 &
   &
  \multicolumn{4}{l}{\textbf{The model fails to choose an answer.}} \\ \cmidrule(l){2-6} 
 &
  \multirow{5}{*}{IAO} &
  Subquestion &
  Input &
  Action &
  Output \\ \cmidrule(l){3-6} 
 &
   &
  \begin{tabular}[c]{@{}l@{}}1. What is the daily passenger traffic \\ at 30th Street Station?\end{tabular} &
  \begin{tabular}[c]{@{}l@{}}We need to find or estimate the number \\ of people passing through 30th Street \\ Station every day.\end{tabular} &
  \begin{tabular}[c]{@{}l@{}}Research or estimate the daily \\ passenger traffic at 30th \\ Street Station.\end{tabular} &
  \begin{tabular}[c]{@{}l@{}}Let's say the daily passenger \\ traffic at 30th Street Station \\ is 100,000 people.\end{tabular} \\
 &
   &
  \begin{tabular}[c]{@{}l@{}}2. What is the seating capacity of \\ Dorton Arena?\end{tabular} &
  \begin{tabular}[c]{@{}l@{}}We need to find the seating capacity \\ of Dorton Arena.\end{tabular} &
  \begin{tabular}[c]{@{}l@{}}Research the seating capacity \\ of Dorton Arena.\end{tabular} &
  \begin{tabular}[c]{@{}l@{}}Let's assume Dorton Arena has \\ a seating capacity of 7,610 people.\end{tabular} \\
 &
   &
  \begin{tabular}[c]{@{}l@{}}3. Can the daily passenger traffic at \\ 30th Street Station fit in Dorton Arena?\end{tabular} &
  \begin{tabular}[c]{@{}l@{}}{[}Output from Step 1{]} and\\  {[}Output from Step 2{]}\end{tabular} &
  \begin{tabular}[c]{@{}l@{}}Compare 100,000 (daily passenger traffic) \\ with 7,610 (seating capacity of Dorton Arena).\end{tabular} &
  100,000 is much greater than 7,610. \\ \cmidrule(l){3-6} 
 &
   &
  \multicolumn{4}{l}{\textbf{Therefore, the answer is No.}} \\ \bottomrule
\end{tabular}%
}
\caption{Output example from StrategyQA dataset. The final results are in \textbf{bold}.}
\label{tab:strategyqa_example}
\end{table*}

Our experimental results confirmed the effectiveness of our proposed prompting approach under the zero-shot setting. We summarize the advantages of our method compared to conventional chain-of-thought method as follows:
(i) \textbf{Clarity and structure}: IAO requires each step of the reasoning process to be explicit making the reasoning process clear and easy to follow. This forces the LLMs to articulate their thoughts and reasoning explicitly. On the other hand, in the conventional zero-shot-CoT steps might be implicit or unclear, requiring more back-and-forth to understand. (ii) \textbf{Focus and Guidance}: IAO subquestions help focus the LLM on specific aspects of the problem, leading to more relevant outputs. Contrarily, zero-shot-CoT can contain open-ended discussions, which may lead to tangent or irrelevant information or can be prone to ambiguity and implicit assumptions. (iii) \textbf{Transparency}: IAO prompting makes the thought process visible, understandable and easily interpretable. In fact, sometimes it occurs that zero-shot-CoT produces a wrong reasoning chain but finds the correct answers. In this case, it is difficult to understand the process that led to that conclusion. (iv) \textbf{Integration with tools}: Open-ended discussions are less suited for integration with external tools (like calculator and python
shell) that can further aid the math computation within
the arithmetic domain \cite{gao2023pal}. On the contrary, the structured format of IAO makes it easier to call external tools to compute and produce results that can successfully be integrated as intermediate outputs in the chain of reasoning.

\subsection{Case studies}

\Cref{tab:aqua_example} compares the solutions of two large language models (LLMs), PALM-2 and GPT-4, to a math word problem from the AQuA dataset. Using the zero-shot-CoT setting, PALM-2 makes an error in the final calculation, resulting in an incorrect answer of \$70. However, it reaches the correct result using the IAO prompt. This is an example of a point raised earlier in the paper discussing the low scores for the AQuA dataset. Arithmetic errors made in early intermediate steps may lead to incorrect conclusions. On the other hand, GPT-4 appears to be less prone to such mistakes. GPT-4 achieved the correct answer of \$78.20 by systematically breaking down the problem, setting up the equation, and performing calculations in both zero-shot CoT and IAO settings. 

\Cref{tab:strategyqa_example} shows an example of a real-world reasoning problem from the StrategyQA dataset. The task requires determining whether the daily passenger traffic at 30th Street Station could fit within the seating capacity of Dorton Arena, with the correct answer being ``No.'' This analysis highlights the potential benefits of IAO in guiding LLMs towards accurate solutions. In the zero-shot-CoT setting, PALM-2 achieved the correct answer, leveraging its access to the relevant information (``120,000 passengers'' and ``5,000 seating capacity''). However, its approach lacked explicit reasoning steps. However, GPT-4 failed to choose an answer without additional information. While its initial analysis outlined the necessary comparison, it couldn't access or process the specific data points. On the other hand, in the case of PALM-2, the reasoning chain doesn't explicitly show IAO intervention, but PALM-2 likely accessed relevant data based on the prompt's context. However, GPT-4 through IAO prompting effectively identified the necessary information (daily passengers, arena capacity), estimated values for the missing data (100,000 passengers, 7,610 capacity) and compared the values, and correctly concluded that the arena wouldn't hold all passengers. Loosely speaking, the IAO prompts provided GPT-4 with a clear roadmap to break down the problem, gather information, and perform the comparison, leading to a well-reasoned, transparent and verifiable solution.
More examples are shown in \Cref{tab:commonsenseqa_example}, \Cref{tab:date_example}, \Cref{tab:letter_example}, \Cref{tab:gsm8k_example} and \Cref{tab:object_example} in the Appendix. 

\subsection{Two-stage IAO prompting}
In all previous experiments, we utilized a single API call per sample. While effective, a critical question remains: can employing a two-stage API call per sample further enhance IAO results? To answer this, we propose a focused investigation. We chose to test this hypothesis on the dataset with the lowest gains in terms of performance and test it using PALM-2.

\Cref{tab:2stage} presents the results of comparing one-stage and two-stage IAO prompting. The table shows the performance of both approaches on two datasets, AQuA and GSM8k, for arithmetic reasoning, as well as the average performance across both datasets.

Overall, two-stage IAO achieved the best average performance (74.2\%), outperforming both one-stage zero-shot CoT (69.1\%) and one-stage IAO (73.1\%). On the individual datasets, two-stage IAO achieved the highest score on GSM8k (82.3\%), and on AQuA (64.5\%).
These results suggest that two-stage IAO prompting can be an effective strategy for improving the performance of language models on various tasks. 

These findings are derived from a two-stage prompt approach where the first stage assists the LLM in breaking down the problem, improving comprehension, and the second stage focuses on extracting the answer. This sequential prompting enhances accuracy but demands more computational resources than single-stage prompting. The choice between them hinges on the trade-off between accuracy and resource availability. For simpler tasks, single-stage prompting may be adequate, while complex tasks justify the extra computational expense of two results in improved accuracy.

\begin{table}[th]
\centering
\resizebox{0.99\columnwidth}{!}{%
\begin{tabular}{@{}llll@{}}
\toprule
                     & AQuA & GSM8k & Average \\ \midrule
One-stage zero-shot CoT & {61.8} & 76.4  & 69.1    \\
One-stage IAO        & 63.9 & 82.3  & 73.1    \\ \midrule
Two-stage IAO        & \textbf{64.5} & \textbf{83.9}  & \textbf{74.2}       \\ \bottomrule
\end{tabular}%
}
\caption{One stage vs two stage prompting comparison.}
\label{tab:2stage}
\vspace{-.08cm}
\end{table}

\subsection{Few-shot extension}
Providing the model a handful of examples substantially improves the reasoning abilities of LLMs. IAO can be paired with methods to automatically generate examples such as Auto-CoT \cite{zhang2022automatic} or \cite{yasunaga2023large} to automatically generate structured IAO-type answers to be provided as demonstrations. We test 3-shot CoT and 3-shot IAO for the AQuA dataset, the most challenging for both LLMs. 3-shot IAO improves the accuracy by 5 points (76.4 from 71.9) for GPT-4 and by 0.8 in PALM-2 (69.2 from 68.4). 

\subsection{Human evaluation}
We asked 120 human crowdworkers from CloudConnect Research questions aimed at gauging the interpretability and transparency of IAO compared to CoT. For doing this, we sample 10 questions that both IAO and CoT responded correctly in terms of accuracy and 10 questions were both answers were wrong. This is done for the most challenging dataset for each reasoning type, hence AQuA, StrategyQA and Date Understanding. For each question, we present the annotators with the IAO and CoT reasoning chains. We ask three questions: (1a) \textit{if the answers are correct, which reasoning text is more useful?} (1b) \textit{if the answers are wrong, which reasoning text do you prefer to spot the reasoning mistake?} (2) \textit{which reasoning text is more transparent? (3) which reasoning text is easier to interpret?} \Cref{tab:human_eval} presents the key findings.
\begin{table}[h]
\centering
\begin{tabular}{@{}lccc@{}}
\toprule
\textbf{Question} & \textbf{AQuA} & \textbf{StrategyQA} & \textbf{Date} \\
\midrule
(1a) Useful         & 12 & 9 & 27 \\
(1b) Efficiency     & 86 & 73 & 87 \\
(2) Transparency    & 64 & 76 & 89 \\
(3) Interpretability & 74 & 71 & 83 \\
\bottomrule
\end{tabular}
\caption{Human evaluation of IAO and CoT reasoning chains. The values reported are percentages of times IAO is chosen over CoT.}
\label{tab:human_eval}
\end{table}
In general, when the answer is correct, IAO is perceived as notably less useful, with only 12\% and 9\% preference, in AQuA and StrategyQA respectively. This is to be expected, as IAO prompts generates longer reasoning chains compared to CoT and favours redundancy of information. However, when the answer is wrong, IAO is much more favored, with preferences being: 86\% for AQuA, 73\% for StrategyQA, and 87\% for Date. This suggests that IAO is considered more efficient compared to CoT for determining errors in the reasoning process.
In general, IAO is preferred in terms of transparency and interpretability of the reasoning text as reported by the third and fourth row, indicating a clearer and understandable presentation in IAO compared to CoT reasoning chains. 